\documentclass{article} % For LaTeX2e
\usepackage[final]{colm2026_conference}

\usepackage{microtype}
\usepackage{hyperref}
\usepackage{url}
\usepackage{booktabs}
\usepackage{lineno}

\definecolor{darkblue}{rgb}{0, 0, 0.5}
\hypersetup{colorlinks=true, citecolor=darkblue, linkcolor=darkblue, urlcolor=darkblue}

\usepackage{amsmath,amsfonts,bm}

\def\secref#1{section~\ref{#1}}
\def\eqref#1{equation~\ref{#1}}
\def\1{\bm{1}}

\DeclareMathAlphabet{\mathsfit}{\encodingdefault}{\sfdefault}{m}{sl}
\SetMathAlphabet{\mathsfit}{bold}{\encodingdefault}{\sfdefault}{bx}{n}

\DeclareMathOperator*{\argmax}{arg\,max}

\usepackage{float}
\usepackage{amssymb}
\usepackage{array}
\usepackage{multirow}
\usepackage{placeins}
\usepackage{makecell}
\usepackage{graphicx}
\usepackage{adjustbox}

\title{LLM Router: Rethinking Routing with Prefill Activations}

\author{%
  \normalfont Tanay Varshney\thanks{Equal contribution.} \quad
  Annie Surla\footnotemark[1] \quad Michelle Xu \quad
  Gomathy Venkata Krishnan \\
  Maximilian Jeblick \quad David Austin \quad Neal Vaidya \quad Davide Onofrio \\
  \makebox[\linewidth][c]{NVIDIA}%
}

\begin{document}

\ifcolmsubmission
\linenumbers
\fi

\maketitle

\begin{abstract}
Existing routers rely on semantic query features or handcrafted features, which
often fail to capture model-specific failures or intrinsic task difficulty. We
instead route using internal LLM activations, specifically the residual stream.
Our key idea, Encoder-Target Decoupling, separates the model
that produces the predictive signal (the Encoder) from the model whose
correctness is being estimated (the Target), allowing open-weight encoders to
predict the performance of closed-source target models. We evaluate layerwise
geometric probes, finding that Fisher Separability ($J$) effectively identifies
informative layers, supported by Effective Dimensionality ($d_{\mathrm{eff}}$)
diagnostics. We then utilize a SharedTrunkNet, a joint multi-output MLP that
predicts simultaneous correctness probabilities across candidate models using
concatenated prefill features. In our experiments, SharedTrunkNet consistently
outperforms semantic baselines. At its best, SharedTrunkNet closes 45.58\% of the
gap between the strongest standalone model and the oracle while achieving
74.31\% cost savings relative to the most expensive model. These results
demonstrate that prefill activations provide a robust routing signal,
establishing activation-based routing as a high-performance alternative to
purely semantic selection.
\end{abstract}

% ================================================================
\section{Introduction}
\label{sec:intro}

The core task of a router is to understand whether an LLM is going to answer a
question correctly and in doing so how much it is going to cost. With these two
inputs, any routing logic can be applied based on application logic. Current
approaches use either a semantic signal, or a set of handcrafted features, as
the core feature to predict LLM correctness. These signals are typically coarse,
and while providing a rough domain specific understanding, lack the fidelity to
judge non surface level complexity and relationships. In this work we contribute

\begin{itemize}
  \item \textbf{Geometrical analysis of LLM's residual stream at prefill as
    feature:} We study Effective Dimensionality ($d_{\mathrm{eff}}$), anisotropy
    ($\alpha$), and Fisher Separability ($J$), and find Fisher $J$ to be the most
    practically useful criterion for identifying separable layers in an LLM's
    residual stream to extract features for routing prediction
  \item \textbf{LLM Encoder-Target decoupling} We show that open-weight encoders
    can serve as strong predictors of closed-source target performance, and in
    several cases, hidden states of a different model outperform the target
    model's own hidden states.
  \item \textbf{SharedTrunkNet:} We utilize a joint multi-output MLP that
    leverages cross-model context to predict simultaneous correctness
    probabilities.
  \item \textbf{Routing evaluation:} We evaluate routing at both the per-model
    and global levels across frontier, small, and mixed model pools.
\end{itemize}
% ================================================================
\section{Related Work}
\label{sec:related}

LLM routing learns a per-query policy that assigns inputs to one of several
candidate models under accuracy, cost, and latency constraints. We organize
prior work by the \emph{decision signal}---what information the router observes
before committing to a model---and identify the gap addressed by
encoder--target decoupling.

\textbf{Cascade and heuristic routing.}
Cascade systems such as FrugalGPT~\citep{chen2023frugalgpt} defer hard queries
to stronger models only after weaker ones fail a quality check. This amortizes
cost on easy inputs, but expected compute grows with the number of stages
traversed, and early rejectors must be calibrated to the same quality metric as
the final model. Cascades therefore buy robustness by spending additional
sequential inference, not by improving the quality of a single routing
decision. We instead commit to one model before generation, and ask whether
richer signals can raise that decision's accuracy without multi-stage fallback.

\textbf{Semantic and black-box routing.}
Most learned routers observe only the query (or a text embedding thereof) and
predict which model will succeed. Preference-trained routers map queries to
model choices from human
comparisons~\citep{ong2024routellm}; retrieval- and graph-based methods
propagate past correctness through embedding neighborhoods or relational
structure~\citep{li2025rethinking,zhang2025avengers,zhu2024graphrouter,feng2025graphrouter};
and encoder-based routers score models with shallow or finetuned text
representations~\citep{xu2025fusionfactory}. Orthogonal formulations cast
routing as item-response estimation~\citep{song2025irtrouter}, causal regret
minimization from observational logs~\citep{tsiourvas2025causal}, or joint
selection of models and reasoning
strategies~\citep{pan2025routetoreason,xu2025fusionfactory}.
These approaches share a common inductive bias: model competence is predictable
from the input distribution alone. That bias fails when semantically similar
queries expose \emph{model-specific} weaknesses that are invisible in the
query text. Semantic routers therefore remain strong black-box baselines, but
leave open whether prefill-time representations carry complementary
information.

\textbf{Mechanistic routing}
A growing line of work shows that LLMs encode their own reliability in hidden
states. Correctness directions form in the residual stream during
prefill~\citep{cencerrado2025noanswer}; internal circuits support
self-prediction of failure~\citep{ghasemabadi2025llms}; and activation
statistics enable inexpensive intent
classification~\citep{chen2025fast}. These results suggest that prefill
activations are a higher-fidelity routing signal than query embeddings.
Existing mechanistic probes, however, are typically \emph{matched}: the model
whose states are read is the model whose correctness is predicted. This
precludes closed-source or API-only targets and couples routing cost to the
target's scale---precisely the setting where black-box semantic routers are
used today.

Our work closes this gap with \emph{encoder--target decoupling}. We extract
prefill-activation features from a small open-source encoder and train a router
that predicts correctness of a separately chosen (possibly closed-source)
target.
This retains the black-box deployment setting of semantic routers while using
the richer prefill signal identified by prior internal-activation analyses---without
requiring access to the target's internals.

\begin{figure}[t]
  \begin{center}
    \includegraphics[width=0.85\linewidth]{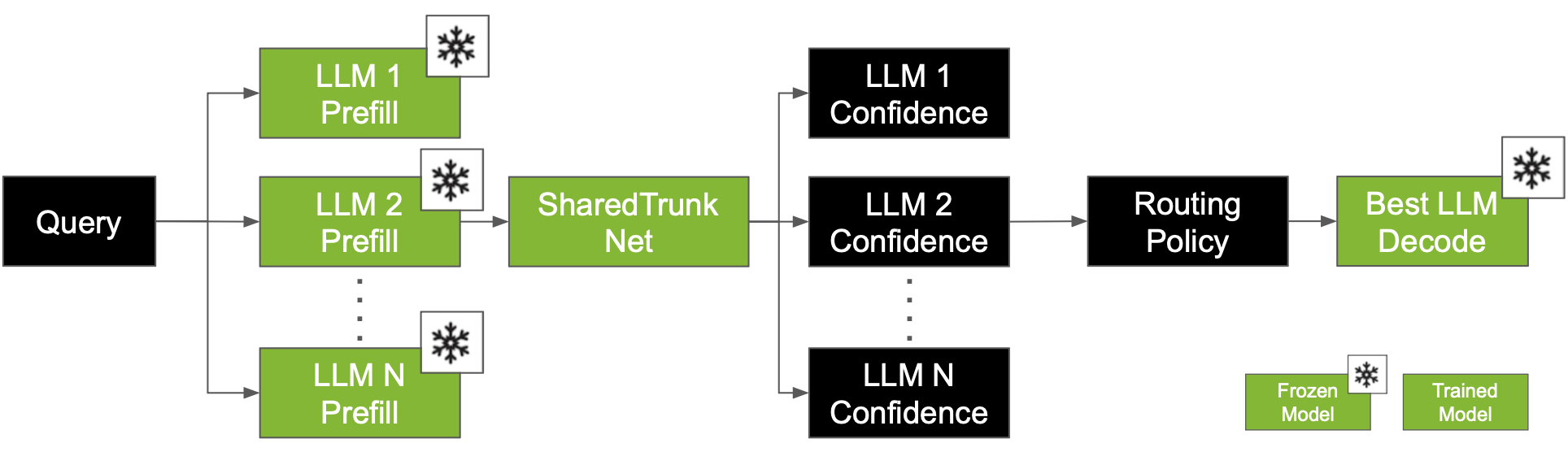}
  \end{center}
  \caption{Overview of the two-stage routing architecture: an Encoder LLM
  extracts prefill features and a SharedTrunkNet estimates per-Target
  correctness probabilities.}
  \label{fig:architecture}
\end{figure}
% ================================================================
\section{Methodology}
\label{sec:methodology}

Figure~\ref{fig:architecture} summarizes our two-stage routing pipeline: an
\textbf{Encoder} LLM produces routing-time features from the query, and a
\textbf{SharedTrunkNet} confidence predictor maps them to a per-\textbf{Target}
correctness estimate $\hat{p}_k(q)$ for each candidate model $k$. Together with
an inference-cost estimate $C_{k,q}$, these two signals drive the routing
decision.

In most cases, inference cost is either API based and a function of input
tokens, output tokens, and per-token pricing, or local-deployment specific and a
function of input tokens, output tokens, inter-token latency, time to first
token, and the cost of renting a GPU per unit time. Because output length is
unknown before generation, the routing decision uses the following estimate of
$C_{k,q}$, with each model's median training output tokens serving as a
verbosity proxy and pricing rates expressed per million tokens:
\begin{equation}
  C_{k,q} = \frac{n_q^{\text{in}}}{10^6} \cdot r_k^{\text{in}} +
              \frac{\tilde{n}_k^{\text{out}}}{10^6} \cdot r_k^{\text{out}}
  \label{eq:cost_est}
\end{equation}
At test time, $\tilde{n}_k^{\text{out}}$ is fixed to model $k$'s median
training OSL, while the observed input length $n_q^{\text{in}}$ varies per
query. This estimate is used only for the pre-generation routing decision.
After the routing choices are fixed, all reported target costs and cost savings
are calculated using the observed input and output token counts for each routed
query. Observed OSL is never provided to the router or used as a
correctness-prediction feature, avoiding ground-truth leakage while ensuring
that output-length variance is reflected in the reported expenditure. The
prefill router also incurs the encoder's own prefill compute, but this overhead
is small: on the frontier pool it adds only $0.76\%$ to the routed dollar cost
and $0.79\%$ to end-to-end latency and does not change the routing comparisons
(Appendix~\ref{sec:appendix_end_to_end_cost},
Tables~\ref{tab:end_to_end_cost_points} and~\ref{tab:end_to_end_latency}).

Given these two signals, each candidate model receives a routing score:
\begin{equation}
  s_{k,q} = \lambda\,\hat{p}_k(q) - (1-\lambda)\,\widetilde{C}_{k,q}
  \label{eq:score}
\end{equation}
where $\widetilde{C}_{k,q} = (C_{k,q} - C_{\min})/(C_{\max} - C_{\min})$ is
cost normalised to $[0,1]$ over the training-set range, and $\lambda \in [0,1]$
is a simple heuristic scalar controlling the accuracy--cost trade-off
($\lambda{=}1$: highest-confidence; $\lambda{=}0$: cheapest; sweeping $\lambda$
traces the full curve). The routing policy selects
$k^*(q) = \argmax_{k \in \mathcal{K}} s_{k,q}$.

\subsection{Prefill router (ours)}
\label{sec:prefill}

Following \citet{chen2025fast}, we propose leveraging an LLM's internal
activations during prefill as a predictive signal for model
correctness---a fundamentally different approach from coarse query
embeddings.

We distinguish between the \emph{Encoder LLM}, which provides hidden-state
signals, and the \emph{Target LLM}, whose performance is predicted. This
decoupling lets us approximate closed-source model capabilities using
open-weights encoders. We evaluate three operational modes: \texttt{per\_model}
(fixed assignments), \texttt{single} (one encoder for all targets), and
\texttt{auto} (optimal encoder per target).

For a query $q$, we extract hidden states from the upper half of the encoder's
transformer layers \citep{lugoloobi2026llms} ($L/2$ to $L$, where $L$ is the
total number of layers), compare last-token vs.\ mean-pooling, and apply PCA to
reduce dimensionality ($d \in \{50, \ldots, 300\}$). We treat the last prefill
token as an operational upper bound on the information available at any single
prefill position: under causal attention, its hidden state can aggregate the
entire prompt, whereas earlier positions cannot incorporate later prompt
tokens. This provides architectural intuition for its predictive utility,
consistent with the pre-generation activation evidence of
\citet{lugoloobi2026llms}, but does not establish that the measured directions
causally determine correctness. Our probes measure predictive association, not
the effect of intervening on a specific representation or circuit. Layer
selection is determined by a grid search validated with 5-fold stratified
cross-validation using $L_2$-regularized logistic regression; these choices
align with the geometric Fisher Separability criterion analysed in
\secref{sec:theory}. The final feature
vector concatenates PCA-reduced features across all $K$ targets, where
$\mathbf{f}_k \in \mathbb{R}^{d_k^{\text{pca}}}$ is the PCA-reduced
hidden-state feature vector for target $k \in \mathcal{K}$:
\begin{equation}
  \mathbf{x} = [\mathbf{f}_1 \mid \mathbf{f}_2 \mid \cdots \mid
  \mathbf{f}_K] \in \mathbb{R}^{\sum_k d_k^{\text{pca}}}
  \label{eq:feature_concat}
\end{equation}

Correctness probabilities are modeled with \textbf{SharedTrunkNet}, a
multi-output MLP mapping $\mathbf{x}$ to simultaneous $P(\text{correct})$
estimates for all $K$ targets in a single forward pass. We train 10
independently seeded instances using BCEWithLogitsLoss and Adam on an 85/15
train/validation split, with early stopping. The top 5 seeds by validation BCE loss are retained; at inference,
their predictions are averaged.
Joint optimization provides cross-model context (features from one encoder
supply evidence about difficulty for other models) and inherent calibration
(outputs remain on a comparable scale).

\subsection{Semantic baselines}
\label{sec:semantic}

As black-box baselines, semantic routers predict per-model correctness
$P(\text{model}_k\text{ correct}\mid q)$ independently of the candidate LLMs,
using features derived exclusively from the input query text. We extract four
feature families per query: (i) dense question embeddings (e.g., llama-nemo-v2
\citep{lee2025nvembed}), optionally PCA-reduced to 128 or 256 dimensions;
(ii) kNN statistics via FAISS \citep{johnson2019faiss} flat-L2 indices over
$K \in \{5, 10, 25, 50, 100\}$ neighbors (neighbor-correct ratio,
distance-weighted correctness, class-conditional distance means);
(iii) handcrafted text-complexity metrics spanning readability (Flesch-Kincaid,
Coleman-Liau, ARI, SMOG), lexical diversity (type-token ratio, hapax legomenon
ratio, Yule's $K$), structural signals (negation count, math/code presence,
parenthetical nesting depth, number density), and information-theoretic measures
(character entropy, zlib compression ratio); and (iv) GPT-2
\citep{radford2019language} signals (mean log-probability, perplexity,
token-level entropy).

These features feed a range of architectures:
\begin{itemize}
  \item \textbf{Rule-based kNN routers:} kNN Majority Vote and kNN Score
    ($\sum_i w_i \cdot y_i$, inverse distance-weighted) \citep{li2025rethinking}.
  \item \textbf{Per-model learned classifiers:} An independent predictor per
    target model---Logistic Regression (LR), XGBoost \citep{chen2016xgboost},
    plain MLP, or a Multi-Task neural network (MT) with joint correctness and
    difficulty heads.
  \item \textbf{Unified Network:} A single shared backbone predicting correctness for all models jointly in one forward pass.
  \item \textbf{End-to-End text models:} LoRA-finetuned \citep{hu2021lora}
    DeBERTa-v3-base \citep{he2021debertav3} bypassing pre-computed embeddings
    entirely, in ``Shared'' (shared LoRA, per-model heads) and ``Per-model''
    (independent encoders) configurations.
\end{itemize}

Training used multi-seed initialization, Adam/AdamW optimizers, and early
stopping, with the best model selected by mean calibration-set AUC. Raw
predictions were passed through a calibration suite
\citep{li2025graphcalibration}: Platt scaling, isotonic regression, percentile
mapping, and $z$-score normalization. We also swept kNN distance-weighting
kernels, FAISS index sizes, and multi-task loss $\lambda$ weights, evaluating
over 1,300 configurations in total across architectures and embedding models.
Additional architectures---IRT/MIRT-2PL \citep{song2025irtrouter}, GNN-based
routing \citep{feng2025graphrouter, zhu2024graphrouter}, LLM-as-judge soft
labels, disagreement-weighted training, and two-stage tiered routing---were
explored but yielded no improvements distinguishable from noise.

% ================================================================
\section{Experimental setup and evaluation protocol}
\label{sec:setup}

\subsection{Correctness labels and evaluation harness}
\label{sec:llm_eval}
In addition to data from LLMRouterBench \citep{li2026llmrouterbench},
we design a lightweight evaluation harness that queries each model across three
benchmarks: MMLU-Pro \citep{wang2024mmluprorobustchallengingmultitask}, Humanity's Last Exam \citep{phan2026hle} (HLE), and LiveCodeBench \citep{jain2024livecodebenchholisticcontaminationfree} (LCB).
Models are queried via OpenAI-compatible streaming endpoints with per-provider
adapters. All models use a maximum of 128{,}000 tokens and their highest
available reasoning effort---\texttt{reasoning\_effort: high} for Claude and
GPT-OSS, and \texttt{enable\_thinking} with provider-recommended sampling for
Qwen and Nemotron. For each (model, question) pair we record the response,
reasoning trace, and wall-clock latency.

Scoring follows standard protocols: exact-match extraction for MMLU-Pro,
GPT-4o-as-a-judge for HLE, and sandboxed pass@1 execution for LCB. All
evaluators produce a unified binary correctness label per (model, question)
pair, which serves as the ground-truth signal for router training and
evaluation. No intentional prompt engineering or harness tuning was applied.

\subsection{Datasets and model pools}
\label{sec:dataset}

We partition queries into three consensus regimes: \textbf{all correct} (every model succeeds), \textbf{all incorrect} (every model fails), and \textbf{model disagreement} (at least one model succeeds while others fail)---the primary regime where routing adds value.

We start with LLMRouterBench, covering 20+ benchmarks across 30 models in two
tiers: a \textbf{Small pool} of 20 models (7B--9B; 13{,}988 entries/model
across 18 benchmarks) and a \textbf{Frontier/Large pool} of 11 models (9{,}662
entries/model across 10 benchmarks). The intersection of these tiers yields
only 2{,}434 entries/model---insufficient to showcase a mixed-tier router.

We therefore collect a \textbf{mixed-tier pool} spanning: Claude Opus 4.6,
OpenAI GPT-5.4, OpenAI GPT-5.2, Qwen 3.5 122B, GPT OSS 120B, Nemotron Super
v3 120B, Nemotron Nano v3 30B, Qwen 3.5 35B, and GPT OSS 20B. This pool
purposely spans different costs and scales. Data is collected across MMLU-Pro,
LiveCodeBench, and HLE using the harness from \secref{sec:llm_eval}, yielding
14{,}469 entries/model. Table~\ref{tab:dataset} summarizes the three pools.

\begin{table}[t]
\begin{center}
\adjustbox{max width=\linewidth}{%
\begin{tabular}{lcccc}
\toprule
\textbf{Dataset} & \textbf{All correct} & \textbf{All fail}
                 & \textbf{Model disagreement} & \textbf{Oracle accuracy} \\
\midrule
Frontier pool & 2,344 (24.26\%) & 1,374 (14.22\%) & 5,944 (61.52\%) & 89.31\% \\
Small pool    & 816 (5.83\%)    & 1,149 (8.21\%)  & 12,023 (85.95\%) & 91.97\% \\
Mixed pool    & 7,596 (52.50\%) & 1,853 (12.81\%) & 5,020 (34.69\%) & 89.35\% \\
\bottomrule
\end{tabular}}
\end{center}
\caption{Dataset statistics. Model disagreement is the primary indicator of
routing opportunity. Oracle accuracy is the accuracy ceiling of the pool.}
\label{tab:dataset}
\end{table}

Data is stratified into an 85--15 train/test split, with calibration experiments using a 75-10-15 train/calibration/test split. Stratifying by both model agreement and benchmark domain ensures proportional task representation and robust in-domain evaluation. All router-training preprocessing and configuration selection are isolated from the test split: scalers and PCA transforms are fit on training data; Fisher statistics and layer selection use training data and labels; and hyperparameter and ensemble-seed selection use only training cross-validation. The resulting transforms, selected layers, and trained predictors are frozen before evaluation on held-out test queries.

\FloatBarrier
\subsection{Evaluation metrics}
\label{sec:eval}

\subsubsection{Level 1: Per-target predictive evaluation}

\textbf{Per-model AUC:} ROC-AUC between predicted $P(\text{correct}_k)$ and
ground-truth labels for each target $k$. Because routing requires judging model
capability across varying difficulty levels, threshold-dependent metrics such
as F1 are unsuitable as primary drivers.

\textbf{Brier score:} Mean squared error between predicted correctness
probabilities and observed binary outcomes, averaged over all queries. It
measures calibration quality per target model, and we report its mean across
targets as our primary calibration metric.

\textbf{Routing delta:} Accuracy and cost on queries routed to vs.\ away from
model $k$.

\subsubsection{Level 2: Global router efficacy (accuracy vs.\ cost)}

We summarize router behavior by sweeping $\lambda \in [0,1]$ and tracing a family of accuracy--cost operating points. To compare routers across pools with different price scales, we evaluate these curves in a normalized inverse-cost space, where higher values correspond to lower mean cost.

\textbf{Padded Area Under the Cost Coverage Curve (P-AUCCC).} Mean cost $\bar{C}$ and accuracy are
normalized to $[0,1]$ using the pool's cheapest/most-expensive models as
anchors:
\begin{align}
  \text{invcost}^{\text{norm}} &=
    \frac{1/\bar{C} - 1/C_{\max}}{1/C_{\min} - 1/C_{\max}} \\
  \text{acc}^{\text{norm}} &=
    \frac{\text{acc} - \text{acc}_{\text{floor}}}
         {\text{acc}_{\text{ceil}} - \text{acc}_{\text{floor}}}
\end{align}
The curve is left-padded at its leftmost accuracy value. P-AUCCC is the
trapezoidal area under this curve ($\in[0,1]$, higher is better).

\textbf{Model Delta Padded AUCCC (MDP-AUCCC).}
Quantifies routing gain over static model selection, where \textbf{P-AUCCC(models)} treats each model as a fixed operating point and serves as the baseline:
\begin{equation}
  \text{MDP-AUCCC} = \text{P-AUCCC}(\text{router}) - \text{P-AUCCC}(\text{models})
  \label{eq:mdp_auccc}
\end{equation}

\textbf{Oracle Distance.} Mean Euclidean distance from each routing-curve point to the oracle corner in
normalised space (lower is better):
\begin{equation}
  D_{\text{oracle}} = \frac{1}{N} \sum_{i}
    \sqrt{
      \left(\Delta\text{invcost}^{\text{norm}}_{i}\right)^2 +
      \left(\Delta\text{acc}^{\text{norm}}_{i}\right)^2
    }
  \label{eq:efficacy}
\end{equation}

% ================================================================
\section{Experimentation \& results}
\label{sec:experiments}

\subsection{Evaluation results}
\label{sec:results}

\subsubsection{Encoder-Target sweeps}

\paragraph{Frontier Pool}
Table~\ref{tab:frontier_sweep} reports per-target AUC for each encoder.
Qwen3.5-122B achieves the highest AUC on every target, suggesting that
large-scale open-weight encoders can serve as especially strong predictors,
including for several closed-source targets. Strikingly, these ``foreign''
open-weight encoders can predict a target's correctness as well as---and for
open targets, better than---signals taken from the target's own hidden states;
we trace this to the encoders' representational geometry (higher effective
dimensionality, isotropy, and Fisher separability) in \secref{sec:theory}.

\textbf{Small Pool} Table~\ref{tab:small_sweep} (Appendix~\ref{sec:appendix_sweeps}) reports
results for the 20-model small pool. Qwen3.5-35B and Qwen3.5-122B again
dominate, achieving the highest AUC on nearly every small-model target.

\textbf{Mixed Pool} Table~\ref{tab:mixed_sweep} (Appendix~\ref{sec:appendix_sweeps}) reports per-target AUC for the mixed-tier
pool, spanning frontier and small models. As with the other pools, Qwen3.5
encoders consistently rank among the strongest predictors across heterogeneous
target architectures.

\begin{table}[t]
\begin{center}
\adjustbox{max width=\linewidth}{%
\begin{tabular}{lcccccccccccc}
\toprule
 & \multicolumn{11}{c}{\textbf{Target}} \\
\cmidrule(l){2-12}
\textbf{Encoder}
  & \multicolumn{1}{c}{\rotatebox{90}{claude-sonnet-4}}
  & \multicolumn{1}{c}{\rotatebox{90}{deepseek-r1}}
  & \multicolumn{1}{c}{\rotatebox{90}{deepseek-v3}}
  & \multicolumn{1}{c}{\rotatebox{90}{gemini-2.5-flash}}
  & \multicolumn{1}{c}{\rotatebox{90}{gemini-2.5-pro}}
  & \multicolumn{1}{c}{\rotatebox{90}{glm-4.6}}
  & \multicolumn{1}{c}{\rotatebox{90}{gpt-5}}
  & \multicolumn{1}{c}{\rotatebox{90}{gpt-5-chat}}
  & \multicolumn{1}{c}{\rotatebox{90}{kimi-k2}}
  & \multicolumn{1}{c}{\rotatebox{90}{openrouter}}
  & \multicolumn{1}{c}{\rotatebox{90}{qwen3-235b}} \\
\midrule
Nemotron-Nano-30B
  & 0.8501 & 0.8154 & 0.7827 & 0.7922 & 0.7350
  & 0.7941 & 0.7872 & 0.7788 & 0.7931 & 0.8092 & 0.7548 \\
gpt-oss-20b
  & 0.8586 & 0.8248 & 0.7910 & 0.7986 & 0.7386
  & 0.8045 & 0.7928 & 0.7899 & 0.8038 & 0.8225 & 0.7551 \\
Qwen3.5-35B
  & 0.9011 & 0.8671 & 0.8333 & 0.8398 & 0.7855
  & 0.8474 & 0.8370 & 0.8366 & 0.8440 & 0.8606 & 0.8135 \\
\textbf{Qwen3.5-122B}
  & \textbf{0.9059} & \textbf{0.8757} & \textbf{0.8427} & \textbf{0.8460}
  & \textbf{0.7978} & \textbf{0.8536} & \textbf{0.8437} & \textbf{0.8488}
  & \textbf{0.8511} & \textbf{0.8620} & \textbf{0.8252} \\
gpt-oss-120b
  & 0.8652 & 0.8324 & 0.8002 & 0.8018 & 0.7475
  & 0.8145 & 0.8036 & 0.7907 & 0.8076 & 0.8239 & 0.7681 \\
Nemotron-Super-120B
  & 0.8991 & 0.8646 & 0.8311 & 0.8343 & 0.7750
  & 0.8418 & 0.8272 & 0.8291 & 0.8409 & 0.8577 & 0.7984 \\
\bottomrule
\end{tabular}}
\end{center}
\caption{Frontier pool: per-target AUC across encoders. Layer selected by
Fisher Separability ($J$); PCA=100, last-token mode. \textbf{Bold} = highest
AUC per target.}
\label{tab:frontier_sweep}
\end{table}

\subsubsection{Per-target predictive evaluation}
\label{sec:per_llm}

To understand how each confidence-predictor backbone generalises
\emph{across} its target models, we aggregate per-model correctness-probability statistics over the
frontier, small, and mixed pools.
For each pool we report: the mean per-model AUC; the mean per-model Brier
score; and the weighted-mean accuracy on queries that were
\emph{routed to} that model, computed under $\argmax$ routing at
$\lambda{=}1$ (see \eqref{eq:score}).
Weights are proportional to query volume.

SharedTrunkNet leads on every metric --- highest AUC, lowest Brier, and
highest routed-to accuracy --- reflecting the benefit of joint multi-target
optimisation and cross-model context.

From the configuration search described in \secref{sec:semantic} (spanning embedding models such as llama-nemo-v2, Qwen-0.6, and DeBERTa), we report one representative configuration per backbone family in Table~\ref{tab:per_llm_routing}, selected as the best-performing variant by mean validation AUC. These benchmarks represent the primary architectural paradigms in recent literature: Unified Multitask and Per-Model Multitask (adapted from MIRT-Router; \citet{song2025irtrouter}), Matrix Factorization (adapted from matrix factorization; \citet{ong2024routellm}), GraphRouter (graph-based router; \citep{feng2025graphrouter}), and kNN (kNN Router; \citet{li2025rethinking})—surfacing the strongest achievable performance within the semantic routing paradigm.

\begin{table}[t]
\centering
\adjustbox{max width=\linewidth}{%
\begin{tabular}{llccc}
\toprule
\textbf{Pool} & \textbf{Architecture}
  & \textbf{Mean per-model AUC} $\uparrow$
  & \textbf{Mean Brier} $\downarrow$
  & \textbf{Wtd.\ Acc (to)} $\uparrow$ \\
\midrule
\multirow{6}{*}{Frontier}
  & SharedTrunkNet [ours]                    & \textbf{0.8560} & \textbf{0.1509} & \textbf{0.7611} \\
  & Unified Multitask [llama-nemo-v2]        & 0.8040 & 0.1756 & 0.7345 \\
  & Matrix Factorization [llama-nemo-v2]     & 0.7943 & 0.1781 & 0.7229 \\
  & Per-Model Multitask [llama-nemo-v2]      & 0.8001 & 0.1874 & 0.6801 \\
  & GraphRouter [Qwen-0.6]                   & 0.7867 & 0.1836 & 0.6978 \\
  & kNN [llama-nemo-v2]                      & 0.7888 & 0.1808 & 0.7100 \\
\midrule
\multirow{6}{*}{Small}
  & SharedTrunkNet [ours]                    & \textbf{0.8260} & \textbf{0.1642} & \textbf{0.7525} \\
  & Unified Multitask [llama-nemo-v2]        & 0.7595 & 0.1912 & 0.7393 \\
  & Matrix Factorization [llama-nemo-v2]     & 0.7485 & 0.1948 & 0.7341 \\
  & Per-Model Multitask [llama-nemo-v2]      & 0.7553 & 0.1925 & 0.7384 \\
  & GraphRouter [Qwen-0.6]                   & 0.7408 & 0.1991 & 0.7055 \\
  & kNN [llama-nemo-v2]                      & 0.7328 & 0.1991 & 0.7341 \\
\midrule
\multirow{6}{*}{Mixed}
  & SharedTrunkNet [ours]                    & \textbf{0.8817} & \textbf{0.1111} & \textbf{0.8336} \\
  & Unified Multitask [llama-nemo-v2]        & 0.8069 & 0.1354 & 0.8304 \\
  & Matrix Factorization [llama-nemo-v2]     & 0.7815 & 0.1411 & 0.8212 \\
  & Per-Model Multitask [llama-nemo-v2]      & 0.8023 & 0.2155 & 0.7627 \\
  & GraphRouter [Qwen-0.6]                   & 0.7936 & 0.1453 & 0.7926 \\
  & kNN [llama-nemo-v2]                      & 0.7729 & 0.1520 & 0.8023 \\
\bottomrule
\end{tabular}}
\caption{Per-backbone aggregated routing metrics for proposed SharedTrunkNet and other Semantic Backbones across the three model pools.
  \emph{Wtd.\ Acc (to)}: weighted-mean accuracy on queries routed to that
  model (router predicted correct). Lower Brier is better; higher AUC and
  Wtd.\ Acc (to) are better.}
\label{tab:per_llm_routing}
\end{table}

\subsubsection{Global Router Evaluation}
\label{sec:global_routing}
\begin{figure}[t]
  \centering
  \begin{minipage}[t]{0.49\linewidth}
    \centering
    {\setlength{\fboxsep}{0pt}\setlength{\fboxrule}{0.2pt}%
    \fbox{\includegraphics[width=\dimexpr\linewidth-2\fboxrule\relax]{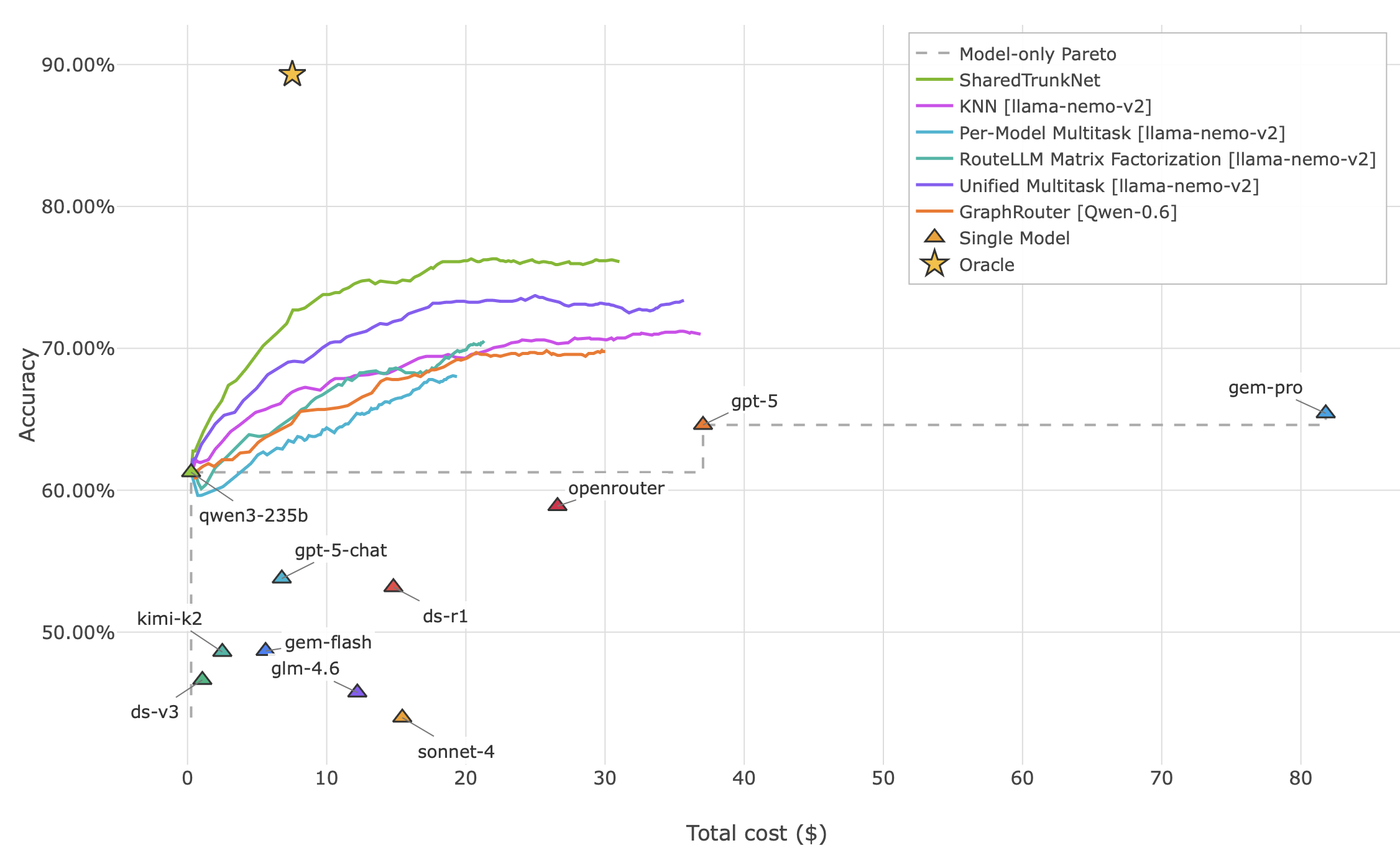}}}
    \caption{Frontier pool: raw accuracy vs.\ total cost (\$). SharedTrunkNet dominates all semantic backbones across the full cost range.}
    \label{fig:frontier_raw}
  \end{minipage}
  \hfill
  \begin{minipage}[t]{0.49\linewidth}
    \centering
    {\setlength{\fboxsep}{0pt}\setlength{\fboxrule}{0.2pt}%
    \fbox{\includegraphics[width=\dimexpr\linewidth-2\fboxrule\relax]{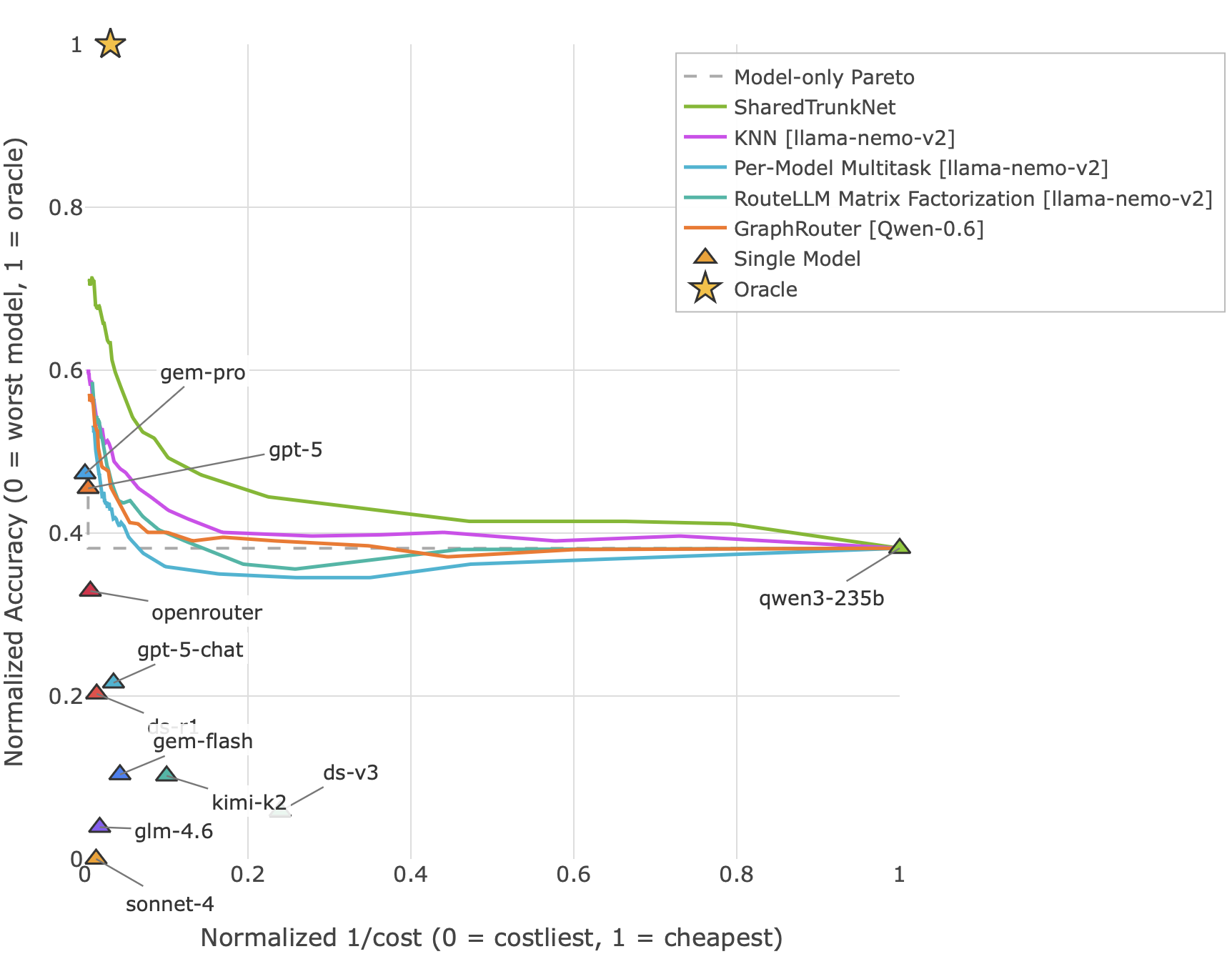}}}
    \caption{Frontier pool: normalized accuracy vs.\ normalized inverse cost. Axes anchored to pool price bounds for cross-pool comparability.}
    \label{fig:frontier_norm}
  \end{minipage}
\end{figure}

To evaluate global routing behavior, we sweep the accuracy--cost trade-off
parameter $\lambda$ and plot the resulting operating points in both raw cost
space (Figure~\ref{fig:frontier_raw}) and normalized inverse-cost space
(Figure~\ref{fig:frontier_norm}). As shown in Table~\ref{tab:headroom_summary},
the gains are largest in the frontier and small pools, where disagreement regimes are broader, and smaller in the mixed pool, where more than half of the queries are answered correctly by all models and the available routing headroom is correspondingly lower. Across all three pools, SharedTrunkNet achieves the strongest overall routing performance among the evaluated methods by P-AUCCC, MDP-AUCCC, and Oracle Distance (Table~\ref{tab:global_routing}), consistently reducing the distance to the theoretical oracle more effectively than the semantic baselines. These results indicate that internal activation geometry provides a stronger
routing signal than the evaluated semantic baselines in our experimental setup.

\begin{table}[t]
\centering
\adjustbox{max width=\linewidth}{%
\begin{tabular}{llccccc}
\toprule
\textbf{Pool} & \textbf{Experiment}
  & \textbf{Oracle Acc.}
  & \textbf{Best Model Acc.}
  & \textbf{Router Acc. Gain} $\uparrow$
  & \textbf{Headroom Captured} $\uparrow$
  & \textbf{Cost Savings} $\uparrow$ \\
\midrule
\multirow{6}{*}{Frontier}
  & SharedTrunkNet [ours]                         & \multirow{6}{*}{89.3\%} & \multirow{6}{*}{65.4\%} & \textbf{$+$10.9pp} & \textbf{45.6\%} & 74.3\% \\
  & Unified Multitask [llama-nemo-v2]             &  &  & $+$8.3pp  & 34.8\%          & 69.5\% \\
  & Matrix Factorization [llama-nemo-v2]          &  &  & $+$5.1pp  & 21.4\%          & 73.9\% \\
  & Per-Model Multitask [llama-nemo-v2]           &  &  & $+$2.7pp  & 11.1\%          & 76.7\% \\
  & GraphRouter [Qwen-0.6]                        &  &  & $+$4.4pp  & 18.5\%          & 68.5\% \\
  & kNN [llama-nemo-v2]                           &  &  & $+$5.8pp  & 24.2\%          & 56.8\% \\
\midrule
\multirow{6}{*}{Small}
  & SharedTrunkNet [ours]                         & \multirow{6}{*}{92.0\%} & \multirow{6}{*}{71.2\%} & \textbf{$+$4.2pp} & \textbf{20.4\%} & 64.2\% \\
  & Unified Multitask [llama-nemo-v2]             &  &  & $+$2.7pp  & 13.1\%          & 49.0\% \\
  & Matrix Factorization [llama-nemo-v2]          &  &  & $-$4.4pp  & $-$21.3\%       & 87.1\% \\
  & Per-Model Multitask [llama-nemo-v2]           &  &  & $+$2.8pp  & 13.3\%          & 53.7\% \\
  & GraphRouter [Qwen-0.6]                        &  &  & $-$0.6pp  & $-$2.7\%        & 65.1\% \\
  & kNN [llama-nemo-v2]                           &  &  & $+$2.5pp  & 12.0\%          & 55.4\% \\
\midrule
\multirow{6}{*}{Mixed}
  & SharedTrunkNet [ours]                         & \multirow{6}{*}{89.4\%} & \multirow{6}{*}{82.2\%} & \textbf{$+$1.2pp} & \textbf{17.3\%} & 29.3\% \\
  & Unified Multitask [llama-nemo-v2]             &  &  & $+$1.1pp  & 15.4\%          & 16.4\% \\
  & Matrix Factorization [llama-nemo-v2]          &  &  & $-$4.5pp  & $-$62.2\%       & 69.6\% \\
  & Per-Model Multitask [llama-nemo-v2]           &  &  & $-$5.9pp  & $-$82.1\%       & 63.8\% \\
  & GraphRouter [Qwen-0.6]                        &  &  & $-$2.9pp  & $-$39.7\%       & 57.4\% \\
  & kNN [llama-nemo-v2]                           &  &  & $-$1.9pp  & $-$26.9\%       & 37.8\% \\
\bottomrule
\end{tabular}}
\caption{Per-pool headroom and cost savings summary. Oracle Acc.\ is the theoretical upper bound; Router Acc.\ Gain is in pp over the best single model; Headroom Captured is the fraction of the oracle--best-model gap closed; Cost Savings is relative to the highest-cost model in the pool and is calculated using observed input and output token counts for each routed query. Pricing is from the OpenRouter API (March 19, 2026).}
\label{tab:headroom_summary}
\end{table}

\begin{table}[t]
\centering
\adjustbox{max width=\linewidth}{%
\begin{tabular}{llcccc}
\toprule
\textbf{Pool} & \textbf{Experiment}
  & \textbf{P-AUCCC} $\uparrow$
  & \textbf{MDP-AUCCC} $\uparrow$
  & \textbf{Oracle Distance $\downarrow$}
  & \textbf{$\Delta$ Oracle Distance} $\uparrow$ \\
\midrule
\multirow{7}{*}{Frontier}
  & Model-only Pareto                            & 0.3817 & ---     & 0.7411 & ---     \\
  & SharedTrunkNet [ours]                        & \textbf{0.4377} & \textbf{+0.0560} & \textbf{0.3437} & \textbf{+0.3973} \\
  & Unified Multitask [llama-nemo-v2]            & 0.4196 & +0.0379 & 0.4102 & +0.3308 \\
  & Matrix Factorization [llama-nemo-v2]         & 0.3846 & +0.0030 & 0.4777 & +0.2633 \\
  & Per-Model Multitask [llama-nemo-v2]          & 0.3682 & -0.0135 & 0.5378 & +0.2033 \\
  & GraphRouter [Qwen-0.6]                       & 0.3887 & +0.0070 & 0.4826 & +0.2585 \\
  & kNN [llama-nemo-v2]                          & 0.4046 & +0.0229 & 0.4614 & +0.2796 \\
\midrule
\multirow{7}{*}{Small}
  & Model-only Pareto                            & 0.3797 & ---     & 0.6187 & ---     \\
  & SharedTrunkNet [ours]                        & \textbf{0.5472} & \textbf{+0.1674} & \textbf{0.3951} & \textbf{+0.2236} \\
  & Unified Multitask [llama-nemo-v2]            & 0.5282 & +0.1485 & 0.4111 & +0.2076 \\
  & Matrix Factorization [llama-nemo-v2]         & 0.4520 & +0.0722 & 0.4910 & +0.1276 \\
  & Per-Model Multitask [llama-nemo-v2]          & 0.5115 & +0.1318 & 0.4132 & +0.2055 \\
  & GraphRouter [Qwen-0.6]                       & 0.4686 & +0.0889 & 0.4581 & +0.1606 \\
  & kNN [llama-nemo-v2]                          & 0.4998 & +0.1201 & 0.4234 & +0.1953 \\
\midrule
\multirow{7}{*}{Mixed}
  & Model-only Pareto                            & 0.1306 & ---     & 0.6807 & ---     \\
  & SharedTrunkNet [ours]                        & \textbf{0.2323} & \textbf{+0.1017} & \textbf{0.3310} & \textbf{+0.3497} \\
  & Unified Multitask [llama-nemo-v2]            & 0.1862 & +0.0556 & 0.3484 & +0.3324 \\
  & Matrix Factorization [llama-nemo-v2]         & 0.1648 & +0.0342 & 0.5539 & +0.1268 \\
  & Per-Model Multitask [llama-nemo-v2]          & 0.1521 & +0.0215 & 0.6059 & +0.0748 \\
  & GraphRouter [Qwen-0.6]                       & 0.1867 & +0.0561 & 0.4982 & +0.1825 \\
  & kNN [llama-nemo-v2]                          & 0.1687 & +0.0381 & 0.4514 & +0.2294 \\
\bottomrule
\end{tabular}}
\caption{Global routing evaluation across pools. P-AUCCC and MDP-AUCCC are higher-is-better; Oracle Distance is lower-is-better; $\Delta$ Oracle Distance is reduction relative to Model-only Pareto. \textbf{Bold} = best router per pool; Model-only Pareto excluded from bolding.}
\label{tab:global_routing}
\end{table}

\subsubsection{Robustness to distribution shift}
\label{sec:ood}
Because a deployed router must handle queries from shifting distributions, we
stress-test SharedTrunkNet under two controlled shifts on the mixed pool (full
evidence in Appendix~\ref{sec:appendix_domain_shift}). Under an
in-distribution category shift---training on MMLU-Pro STEM and evaluating on
held-out Humanities and Social Sciences---SharedTrunkNet loses only $0.74$
accuracy points relative to its IID reference, the smallest degradation among
all routers (Table~\ref{tab:mmlupro_shift}). Under the harder cross-distribution
test---removing all HLE questions from training---argmax accuracy drops more
sharply ($-11.43$ points), yet SharedTrunkNet still retains the highest ranking
quality (macro AUC $0.6367$ versus at most $0.5541$ for the semantic baselines;
Table~\ref{tab:hle_shift}). The prefill signal therefore transfers well across
related domains and degrades gracefully---rather than catastrophically---under
severe shift.

\subsubsection{Overfitting controls and statistical reliability}
\label{sec:reliability}
Because prefill features are high-dimensional, SharedTrunkNet is, in principle,
the method most exposed to overfitting on the in-domain split. We control for
this in three ways---PCA compression of each encoder's hidden states, a single
Fisher-selected layer validated by 5-fold cross-validation
(\secref{sec:prefill}), and a five-seed ensemble with early stopping---and then
verify that the reported gains survive statistical scrutiny. A query-level
nonparametric bootstrap ($N{=}1000$) yields tight 95\% confidence intervals: the
largest one-sided deviation from a SharedTrunkNet point estimate across all three
pools is only $0.0135$ AUC, $0.0072$ Brier, and $0.0417$ P-AUCCC, and
training-seed variation is comparably small
(Appendix~\ref{sec:appendix_uncertainty}). A \emph{paired} bootstrap against the
strongest semantic baseline (Unified Multitask) confirms the advantage is
significant for 10 of the 12 pool--metric comparisons, including P-AUCCC in all
three pools (Appendix~\ref{sec:appendix_paired_bootstrap},
Table~\ref{tab:paired_bootstrap_unified}). A forced routing-ratio sweep that
isolates ranking quality from cost-aware $\lambda$ tuning shows SharedTrunkNet
leading the semantic baselines throughout the practically relevant low-ratio
regime (Appendix~\ref{sec:appendix_routing_ratio},
Figure~\ref{fig:routing_ratio_sweeps}), and the layer-selection ablation
(Appendix~\ref{sec:appendix_layer_selection},
Table~\ref{tab:layer_selector_ablation}) shows the chosen Fisher layers track the
exhaustive per-target oracle rather than a split-specific optimum. Together with
the domain-shift results above, these analyses indicate the in-domain gains
reflect a reproducible signal rather than overfitting.

% ================================================================
\section{Geometric diagnostics of prefill signals}
\label{sec:theory}

A significant empirical finding is that ``foreign'' encoders (e.g.,
Qwen-35B/122B) consistently outperform a target model's own internal states
in predicting its correctness. To characterize this predictive association,
we evaluate encoder hidden states across three geometric dimensions.

We hypothesize that a robust predictive signal requires high dimensionality,
isotropy, and linear separability, quantified as follows:

\begin{itemize}
  \item \textbf{Effective dimensionality ($d_{\text{eff}}$):} Participation
    ratio of covariance eigenvalues $\sigma_i$:
    \begin{equation}
      d_{\text{eff}} = \frac{\left(\sum_i \sigma_i\right)^2}{\sum_i \sigma_i^2}
      \label{eq:deff}
    \end{equation}
    Higher $d_{\text{eff}}$ indicates that information is widely distributed,
    reducing PCA information loss.

  \item \textbf{Representational anisotropy ($\alpha$):} Pairwise cosine
    similarity between hidden-state vectors:
    \begin{equation}
      \alpha = \frac{2}{n(n-1)} \sum_{i<j} \cos(\mathbf{h}_i, \mathbf{h}_j)
      \label{eq:anisotropy}
    \end{equation}
    Lower $\alpha$ (higher isotropy) prevents the ``narrow cone'' pathology
    where outlier dimensions dominate and collapse class-discriminative
    differences.

  \item \textbf{Fisher Separability ($J$):} Multivariate Fisher criterion on
    PCA-reduced features:
    \begin{equation}
      J = \frac{\|\boldsymbol{\mu}_1 - \boldsymbol{\mu}_0\|^2}
               {\mathrm{tr}(\Sigma_0) + \mathrm{tr}(\Sigma_1)}
      \label{eq:fisher}
    \end{equation}
    where class 1 and 0 represent correct and incorrect responses, respectively. Unlike
    $d_{\text{eff}}$ and $\alpha$, which are label-free preconditions, $J$ is a
    target-dependent measure of task-specific separability.
\end{itemize}

These statistics describe geometric associations between activations and
correctness labels. They do not identify a causal mechanism or show that an
intervention on the measured features would change model correctness.

Fisher $J$-based layer selection closely matches the trends observed in
empirical probe sweeps, providing an interpretable and efficient heuristic for
identifying separable layers. A head-to-head layer-selection ablation confirms
this: Fisher $J$ is the best or tied-best practical selector for 16 of the 18
pool--encoder combinations (and within $0.0003$ AUC of the best selector in the
other two), while staying close to the exhaustive per-target layer oracle
(Appendix~\ref{sec:appendix_layer_selection},
Table~\ref{tab:layer_selector_ablation}). While $d_{\mathrm{eff}}$ and anisotropy
often peak at different layers, those peaks do not consistently correspond to the
strongest target-level separability. Concatenating layers selected independently
by these metrics yielded no material AUC improvement.

% ================================================================
\section{Conclusion}
\label{sec:conclusion}

We presented an activation-based approach to LLM routing rather than semantic query features. Our central idea, Encoder-Target Decoupling, allows open-weight encoders to estimate the correctness of both open and closed-source target models, and our SharedTrunkNet architecture leverages joint multi-target prediction to improve per-model confidence quality and global routing performance.

Across frontier, small, and mixed model pools, SharedTrunkNet consistently outperforms the evaluated semantic baselines, with the largest gains appearing in the frontier pool. Geometrically, we find that Fisher Separability ($J$) is the most practically useful probe for selecting informative layers, while Effective Dimensionality and anisotropy provide complementary descriptive diagnostics.

At the same time, our activation analyses are correlational: they establish predictive utility but do not identify a causal mechanism or show that intervening on a representation changes correctness. Because output length is unavailable before generation, the routing decision estimates it using each model's median training output length; however, all reported costs and cost savings are calculated afterward using the observed input and output token counts for each routed query. Our global routing evaluation also depends on a normalized inverse-cost metric suite that merits further robustness analysis. Subject to these limitations, the results suggest that prefill activations provide a strong routing signal and that activation-based routing is a promising alternative to purely semantic LLM selection.

In the frontier pool, SharedTrunkNet closed \textbf{45.58\%} of the accuracy gap between the strongest standalone model and the theoretical oracle, while achieving \textbf{74.31\%} cost savings relative to the highest-cost model, 53.62\% lower Oracle Distance, and a 14.67\% increase in P-AUCCC over the model-only Pareto frontier, confirming that activation-based routing provides a robust and cost-effective foundation for collaborative LLM systems.

% ================================================================
\newpage
\bibliography{llm_router_paper_submission}
\bibliographystyle{colm2026_conference}

% ================================================================
\clearpage
\appendix

% ----------------------------------------------------------------
\section{Layer-Selection Ablation}
\label{sec:appendix_layer_selection}

We compare layer-selection strategies over the top half of the available
transformer layers for every encoder and model pool. For each encoder, the
entries in Table~\ref{tab:layer_selector_ablation} are mean five-fold
stratified-CV logistic-regression probe AUCs across the pool's routing targets,
using PCA-100 last-token features. Random is the expected AUC under a uniform
choice among the probed layers; Last uses the final layer; $d_{\mathrm{eff}}$
selects the layer of maximum effective dimensionality; $\alpha$ selects the
least anisotropic layer; and Fisher $J$ selects the layer of maximum Fisher
separability. BestLR is the exhaustive per-target layer oracle and is included
only as an upper-bound reference.

\begin{table}[H]
\begin{center}
\small
\setlength{\tabcolsep}{4.5pt}
\adjustbox{max width=\textwidth}{%
\begin{tabular}{llrrrrrr}
\toprule
\textbf{Pool} & \textbf{Encoder} & \textbf{Random} & \textbf{Last}
  & $\boldsymbol{d}_{\mathbf{eff}}$ & $\boldsymbol{\alpha}$
  & \textbf{Fisher $J$} & \textbf{BestLR} \\
\midrule
Frontier & Nemo-Nano-30B  & 0.7871 & 0.7868 & 0.7877 & 0.7890 & \textbf{0.7902} & 0.7905 \\
Frontier & GPT-OSS-20B    & 0.7932 & 0.7876 & 0.7946 & 0.7962 & \textbf{0.7982} & 0.7989 \\
Frontier & Qwen3.5-35B    & 0.8384 & 0.8361 & 0.8392 & 0.8364 & \textbf{0.8424} & 0.8425 \\
Frontier & Qwen3.5-122B   & 0.8336 & 0.8436 & 0.8499 & 0.8466 & \textbf{0.8502} & 0.8504 \\
Frontier & GPT-OSS-120B   & 0.7963 & 0.7961 & 0.8016 & \textbf{0.8050} & \textbf{0.8050} & 0.8050 \\
Frontier & Nemo-Super-120B & 0.8339 & 0.8357 & 0.8355 & 0.8352 & \textbf{0.8363} & 0.8368 \\
\midrule
Small & Nemo-Nano-30B     & 0.7875 & 0.7861 & 0.7889 & 0.7868 & \textbf{0.7902} & 0.7906 \\
Small & GPT-OSS-20B       & 0.7677 & 0.7640 & 0.7676 & 0.7702 & \textbf{0.7722} & 0.7724 \\
Small & Qwen3.5-35B       & 0.8038 & 0.8004 & 0.8082 & 0.8011 & \textbf{0.8094} & 0.8097 \\
Small & Qwen3.5-122B      & 0.7913 & 0.8054 & 0.8055 & 0.8064 & \textbf{0.8092} & 0.8094 \\
Small & GPT-OSS-120B      & 0.7600 & 0.7523 & 0.7607 & 0.7696 & \textbf{0.7697} & 0.7697 \\
Small & Nemo-Super-120B   & 0.7966 & 0.7991 & 0.7986 & 0.7967 & \textbf{0.7998} & 0.8003 \\
\midrule
Mixed & Nemo-Nano-30B     & 0.7441 & 0.7429 & \textbf{0.7483} & 0.7426 & 0.7480 & 0.7485 \\
Mixed & GPT-OSS-20B       & 0.8182 & 0.8164 & 0.8224 & 0.8150 & \textbf{0.8234} & 0.8245 \\
Mixed & Qwen3.5-35B       & 0.8629 & 0.8592 & 0.8533 & \textbf{0.8695} & 0.8693 & 0.8697 \\
Mixed & Qwen3.5-122B      & 0.8529 & 0.8676 & 0.8408 & 0.8696 & \textbf{0.8734} & 0.8734 \\
Mixed & GPT-OSS-120B      & 0.8158 & 0.8052 & 0.8105 & 0.8205 & \textbf{0.8216} & 0.8221 \\
Mixed & Nemo-Super-120B   & 0.8471 & 0.8438 & 0.8493 & 0.8479 & \textbf{0.8521} & 0.8526 \\
\bottomrule
\end{tabular}}
\end{center}
\caption{Layer-selection ablation across all model pools and encoders. Values
are mean AUC across the routing targets in the corresponding pool. Bold marks
the best non-oracle selector in each row; BestLR is the exhaustive per-target
LR layer oracle.}
\label{tab:layer_selector_ablation}
\end{table}

Fisher is the best or tied-best practical selector for 16 of the 18
pool--encoder combinations and remains within 0.0003 AUC of the best selector
in the other two cases. Moreover, its gap to the exhaustive BestLR oracle is
small throughout, supporting Fisher separability as a stable layer-selection
criterion rather than a brittle choice tied to one encoder or pool.

\FloatBarrier
\clearpage

% ----------------------------------------------------------------
\section{Domain-Shift Evaluation}
\label{sec:appendix_domain_shift}

Our router, like the target LLMs it routes, learns regularities in the
training distribution. It is therefore intended to transfer across related
domains represented during training, rather than to provide zero-shot
guarantees for radically different query distributions or previously unseen
model pools. We quantify this operating envelope on the nine-target mixed
pool with two controlled shifts.

\subsection{MMLU-Pro Category Shift}

For an in-distribution category shift, we train on 4,722 MMLU-Pro STEM
questions and evaluate on 3,503 held-out Humanities and Social Sciences
questions. The IID reference trains on the full mixed-pool training set and is
evaluated on exactly the same held-out questions. Table~\ref{tab:mmlupro_shift}
reports argmax-routing accuracy; $\Delta$ is STEM-trained minus IID.

\begin{table}[H]
\begin{center}
\small
\setlength{\tabcolsep}{7pt}
\begin{tabular}{lrrr}
\toprule
\textbf{Router} & \textbf{STEM-trained} & \textbf{IID} & $\boldsymbol{\Delta}$ \textbf{(pp)} \\
\midrule
SharedTrunkNet (Fisher) & \textbf{88.15} & 88.90 & $-0.74$ \\
Per-Model Multitask     & 86.53 & 86.93 & $-0.40$ \\
Unified Multitask       & 87.24 & 88.50 & $-1.26$ \\
Matrix Factorization    & 81.87 & 85.98 & $-4.11$ \\
kNN voting ($k{=}25$)   & 85.36 & 85.78 & $-0.43$ \\
GraphRouter             & 85.61 & 91.29 & $-5.68$ \\
\bottomrule
\end{tabular}
\end{center}
\caption{MMLU-Pro category-shift results on 3,503 held-out Humanities and
Social Sciences questions. Accuracy values and differences are percentage
points. SharedTrunkNet loses only 0.74 points relative to its IID reference.}
\label{tab:mmlupro_shift}
\end{table}

\subsection{Held-Out HLE Stress Test}

For the stronger cross-distribution test, we remove all HLE questions from
training and evaluate only on the 1,225 held-out HLE questions. The IID
reference includes HLE in its training distribution. Along with argmax
accuracy, Table~\ref{tab:hle_shift} reports the macro mean per-target ROC-AUC
under HLE exclusion.

\begin{table}[H]
\begin{center}
\small
\setlength{\tabcolsep}{5.5pt}
\begin{tabular}{lrrrr}
\toprule
\textbf{Router} & \makecell{\textbf{HLE held-out}\\\textbf{accuracy}}
  & \makecell{\textbf{IID}\\\textbf{accuracy}}
  & $\boldsymbol{\Delta}$ \textbf{(pp)}
  & \makecell{\textbf{Held-out}\\\textbf{macro AUC}} \\
\midrule
SharedTrunkNet (Fisher) & \textbf{32.90} & 44.33 & $-11.43$ & \textbf{0.6367} \\
Per-Model Multitask     & 12.82 & 39.02 & $-26.20$ & 0.5541 \\
Unified Multitask       & 30.86 & 42.94 & $-12.08$ & 0.5478 \\
Matrix Factorization    & 19.02 & 41.63 & $-22.61$ & 0.5413 \\
kNN voting ($k{=}25$)   & 23.43 & 29.80 & $-6.37$  & 0.5419 \\
GraphRouter             & \textbf{32.90} & 49.96 & $-17.06$ & 0.5479 \\
\bottomrule
\end{tabular}
\end{center}
\caption{Cross-distribution generalization on HLE. Accuracy values and
differences are percentage points. SharedTrunkNet retains the highest ranking
quality, with macro AUC 0.6367 versus at most 0.5541 for the semantic
baselines.}
\label{tab:hle_shift}
\end{table}

The contrast between the 0.74-point MMLU-Pro category-shift loss and the
11.43-point HLE loss is approximately fifteen-fold. Thus, the prefill signal
transfers well across domain families represented by the broader training
distribution, but degrades when extrapolating from exam-style questions to a
held-out research-grade distribution. Under that stress test,
SharedTrunkNet's accuracy degradation is close to Unified Multitask's, while
its ranking AUC remains stronger.

\FloatBarrier
\clearpage

% ----------------------------------------------------------------
\section{Encoder Target Sweeps}
\label{sec:appendix_sweeps}

\subsection{Small Pool}

Table~\ref{tab:small_sweep} reports per-target AUC across encoder choices for the small pool.

\begin{table}[t]
\begin{center}
\small
\setlength{\tabcolsep}{4pt}
\begin{tabular}{lcccccc}
\toprule
\textbf{Model}
  & \makecell{Nemo\\Nano\\30B}
  & \makecell{gpt\\oss\\20b}
  & \makecell{Qwen\\3.5\\35B}
  & \makecell{Qwen\\3.5\\122B}
  & \makecell{gpt\\oss\\120b}
  & \makecell{Nemo\\Super\\120B} \\
\midrule
DeepHermes-3-8B  & 0.7744 & 0.7550 & 0.7926 & \textbf{0.7928} & 0.7527 & 0.7823 \\
DS-R1-Qwen3-8B   & 0.8047 & 0.7853 & \textbf{0.8290} & 0.8260 & 0.7848 & 0.8140 \\
DS-R1D-Qwen-7B   & 0.7817 & 0.7563 & \textbf{0.8002} & 0.7991 & 0.7552 & 0.7918 \\
Fin-R1           & 0.8143 & 0.7981 & 0.8277 & \textbf{0.8292} & 0.7950 & 0.8215 \\
GLM-Z1-9B        & 0.7923 & 0.7706 & \textbf{0.8141} & 0.8113 & 0.7619 & 0.8020 \\
Intern-S1-mini   & 0.7776 & 0.7610 & \textbf{0.8036} & 0.8012 & 0.7594 & 0.7917 \\
Llama-8B-Inst    & 0.7884 & 0.7701 & 0.8064 & \textbf{0.8085} & 0.7659 & 0.7974 \\
Llama-8B-Med     & 0.7908 & 0.7742 & 0.8066 & \textbf{0.8068} & 0.7691 & 0.8019 \\
Llama-Nemo-8B    & 0.7626 & 0.7470 & \textbf{0.7803} & 0.7795 & 0.7409 & 0.7753 \\
MiMo-7B-RL       & 0.8462 & 0.8222 & \textbf{0.8667} & 0.8632 & 0.8231 & 0.8542 \\
MiniCPM-4.1-8B   & 0.7957 & 0.7798 & \textbf{0.8153} & 0.8136 & 0.7789 & 0.8030 \\
Nemo-Nano-9B     & 0.7937 & 0.7708 & 0.8165 & \textbf{0.8175} & 0.7728 & 0.8022 \\
OpenThinker-3-7B & 0.7911 & 0.7766 & \textbf{0.8014} & 0.8006 & 0.7757 & 0.7972 \\
Qwen2.5-Coder-7B & 0.7783 & 0.7584 & \textbf{0.7954} & 0.7939 & 0.7568 & 0.7870 \\
Qwen3-8B         & 0.7696 & 0.7534 & 0.7936 & \textbf{0.7956} & 0.7534 & 0.7822 \\
cogito-8B        & 0.7782 & 0.7655 & \textbf{0.8026} & \textbf{0.8026} & 0.7593 & 0.7903 \\
gemma2-9b        & 0.7997 & 0.7792 & 0.8212 & \textbf{0.8217} & 0.7792 & 0.8115 \\
glm-4-9b         & 0.8053 & 0.7914 & \textbf{0.8220} & 0.8219 & 0.7885 & 0.8122 \\
granite-3.3-8b   & 0.7795 & 0.7638 & 0.7927 & \textbf{0.7971} & 0.7627 & 0.7896 \\
intlm3-8b        & 0.7797 & 0.7643 & 0.8000 & \textbf{0.8015} & 0.7589 & 0.7895 \\
\bottomrule
\end{tabular}
\end{center}
\caption{Small pool: per-target AUC across encoders. Layer selected by Fisher
Separability ($J$); PCA=100, last-token mode. \textbf{Bold} = highest AUC per target.}
\label{tab:small_sweep}
\end{table}

\FloatBarrier
\clearpage
\subsection{Mixed Pool}

Table~\ref{tab:mixed_sweep} reports per-target AUC across encoder choices for the mixed pool.

\begin{table}[t]
\begin{center}
\adjustbox{max width=\linewidth}{%
\begin{tabular}{lcccccccccc}
\toprule
\textbf{Encoder}
  & \multicolumn{1}{c}{\rotatebox{90}{Claude Opus 4.6}}
  & \multicolumn{1}{c}{\rotatebox{90}{GPT-5.2}}
  & \multicolumn{1}{c}{\rotatebox{90}{GPT-5.4}}
  & \multicolumn{1}{c}{\rotatebox{90}{GPT-OSS 120B}}
  & \multicolumn{1}{c}{\rotatebox{90}{GPT-OSS 20B}}
  & \multicolumn{1}{c}{\rotatebox{90}{Nemotron Nano 30B}}
  & \multicolumn{1}{c}{\rotatebox{90}{Nemotron Super 120B}}
  & \multicolumn{1}{c}{\rotatebox{90}{Qwen 3.5 122B}}
  & \multicolumn{1}{c}{\rotatebox{90}{Qwen 3.5 35B}} \\
\midrule
Nemotron-Nano-30B   & 0.7569 & 0.7366 & 0.7345 & 0.7357 & 0.7458 & 0.7502 & 0.7487 & 0.7592 & 0.7648 \\
gpt-oss-20b         & 0.8313 & 0.8050 & 0.8017 & 0.8063 & 0.8264 & 0.8316 & 0.8228 & 0.8404 & 0.8452 \\
Qwen3.5-35B         & 0.8636 & 0.8545 & 0.8437 & 0.8562 & 0.8789 & 0.8817 & 0.8724 & 0.8825 & 0.8898 \\
\textbf{Qwen3.5-122B} & \textbf{0.8644} & \textbf{0.8590} & \textbf{0.8516} & \textbf{0.8616} & \textbf{0.8859} & \textbf{0.8828} & \textbf{0.8769} & \textbf{0.8858} & \textbf{0.8924} \\
gpt-oss-120b        & 0.8260 & 0.7994 & 0.7990 & 0.8039 & 0.8305 & 0.8317 & 0.8233 & 0.8382 & 0.8428 \\
Nemotron-Super-120B & 0.8470 & 0.8343 & 0.8265 & 0.8378 & 0.8672 & 0.8640 & 0.8563 & 0.8626 & 0.8737 \\
\bottomrule
\end{tabular}}
\end{center}
\caption{Mixed pool: per-target AUC across encoders. Layer selected by Fisher
Separability ($J$); PCA=100, last-token mode. \textbf{Bold} = highest AUC per target.}
\label{tab:mixed_sweep}
\end{table}

% ----------------------------------------------------------------
\clearpage
\section{Top-3 and Bottom-3 Experiments per-Model AUC and Brier Score}
\label{sec:appendix_top3}

\subsection{Frontier Pool}

Tables~\ref{tab:frontier_auc} and~\ref{tab:frontier_brier} show the top-3 and bottom-3 experiments by AUC and Brier score for the frontier pool.

\begin{table}[t]
\begin{center}
\adjustbox{max width=\linewidth}{%
\begin{tabular}{lcccccccccccc}
\toprule
\textbf{Experiment} & \textbf{Mean} & sonnet-4 & ds-r1 & ds-v3
  & g-flash & g-pro & glm-4.6 & gpt-5 & gpt-5c & kimi & openrt & q-235b \\
\midrule
\#1 Unified Multitask [llama-nemo-v2]
  & 0.8040 & 0.8653 & 0.8385 & 0.8049 & 0.8137 & 0.7585
  & 0.7940 & 0.7917 & 0.7804 & 0.7991 & 0.8263 & 0.7714 \\
\#2 Per-Model Multitask [llama-nemo-v2]
  & 0.8001 & 0.8633 & 0.8363 & 0.8024 & 0.8099 & 0.7522
  & 0.7880 & 0.7875 & 0.7791 & 0.7924 & 0.8196 & 0.7703 \\
\#3 Per-Model Multitask [qwen-0.6]
  & 0.7984 & 0.8553 & 0.8329 & 0.8134 & 0.8081 & 0.7335
  & 0.7875 & 0.7861 & 0.7791 & 0.7969 & 0.8190 & 0.7702 \\
\midrule
\#(last-2) GraphRouter [deberta]
  & 0.7669 & 0.8316 & 0.8120 & 0.7798 & 0.7719 & 0.6963
  & 0.7540 & 0.7634 & 0.7457 & 0.7618 & 0.7924 & 0.7269 \\
\#(last-1) kNN [deberta]
  & 0.7692 & 0.8292 & 0.8103 & 0.7784 & 0.7902 & 0.7087
  & 0.7570 & 0.7711 & 0.7508 & 0.7708 & 0.7670 & 0.7281 \\
\#(last) IRT [mpnet]
  & 0.7561 & 0.8051 & 0.7989 & 0.7713 & 0.7789 & 0.7206
  & 0.7642 & 0.7576 & 0.7401 & 0.7756 & 0.6956 & 0.7097 \\
\bottomrule
\end{tabular}}
\end{center}
\caption{Frontier pool --- top 3 and bottom 3 by mean per-model AUC.}
\label{tab:frontier_auc}
\end{table}

\begin{table}[t]
\begin{center}
\adjustbox{max width=\linewidth}{%
\begin{tabular}{lcccccccccccc}
\toprule
\textbf{Experiment} & \textbf{Mean} & sonnet-4 & ds-r1 & ds-v3
  & g-flash & g-pro & glm-4.6 & gpt-5 & gpt-5c & kimi & openrt & q-235b \\
\midrule
\#1 Unified Multitask [llama-nemo-v2]
  & 0.1756 & 0.1459 & 0.1593 & 0.1803 & 0.1758 & 0.1855
  & 0.1830 & 0.1746 & 0.1919 & 0.1838 & 0.1640 & 0.1873 \\
\#2 Per-Model Multitask [qwen-0.6]
  & 0.1770 & 0.1517 & 0.1616 & 0.1771 & 0.1786 & 0.1946
  & 0.1816 & 0.1767 & 0.1901 & 0.1826 & 0.1671 & 0.1857 \\
\#3 Per-Model Multitask [llama-nemo-v2]
  & 0.1874 & 0.1548 & 0.1661 & 0.1826 & 0.1788 & 0.2205
  & 0.1899 & 0.2022 & 0.1944 & 0.1867 & 0.1785 & 0.2068 \\
\midrule
\#(last-2) GraphRouter [deberta]
  & 0.1960 & 0.1734 & 0.1802 & 0.1986 & 0.2001 & 0.2061
  & 0.2076 & 0.1871 & 0.2066 & 0.2058 & 0.1863 & 0.2036 \\
\#(last-1) kNN [deberta]
  & 0.1882 & 0.1624 & 0.1687 & 0.1912 & 0.1857 & 0.1994
  & 0.1938 & 0.1822 & 0.2010 & 0.1917 & 0.1914 & 0.2022 \\
\#(last) IRT [mpnet]
  & 0.1939 & 0.1727 & 0.1793 & 0.1948 & 0.1899 & 0.1969
  & 0.1951 & 0.1858 & 0.2042 & 0.1918 & 0.2137 & 0.2084 \\
\bottomrule
\end{tabular}}
\end{center}
\caption{Frontier pool --- top 3 and bottom 3 by mean per-model Brier score (lower is better).}
\label{tab:frontier_brier}
\end{table}

\FloatBarrier
\clearpage
\subsection{Small Pool}

Tables~\ref{tab:small_auc} and~\ref{tab:small_brier} show the top-3 and bottom-3 experiments by AUC and Brier score for the small pool.

\begin{table}[t]
\begin{center}
\adjustbox{max width=\linewidth}{%
\begin{tabular}{lccccccc}
\toprule
\textbf{Model}
  & \makecell{\#1 Unified\\ Multitask [deberta]}
  & \makecell{\#2 Unified\\ Multitask [qwen-0.6]}
  & \makecell{\#3 Unified\\ Multitask [llama-nemo-v2]}
  & \makecell{last kNN\\{} [mpnet]}
  & \makecell{last-1 kNN\\{} [deberta]}
  & \makecell{last-2 Finetuned\\ deberta v3} \\
\midrule
DeepHermes-3-8B  & 0.7313 & 0.7348 & \textbf{0.7390} & 0.6837 & 0.6892 & 0.7057 \\
DS-R1-Qwen3-8B   & \textbf{0.7770} & 0.7628 & 0.7709 & 0.7200 & 0.7150 & 0.6995 \\
DS-R1D-Qwen-7B   & 0.7455 & \textbf{0.7553} & 0.7532 & 0.7082 & 0.7001 & 0.6713 \\
Fin-R1           & 0.7881 & 0.7870 & \textbf{0.7932} & 0.7372 & 0.7507 & 0.7527 \\
GLM-Z1-9B        & 0.7662 & 0.7657 & \textbf{0.7698} & 0.7198 & 0.7198 & 0.7009 \\
Intern-S1-mini   & 0.7570 & \textbf{0.7580} & 0.7482 & 0.7185 & 0.6856 & 0.6886 \\
Llama-8B-Inst    & \textbf{0.7668} & 0.7602 & 0.7552 & 0.7121 & 0.7117 & 0.7246 \\
Llama-8B-Med     & \textbf{0.7624} & 0.7614 & 0.7538 & 0.7174 & 0.7299 & 0.7087 \\
Llama-Nemo-8B    & \textbf{0.7463} & 0.7370 & 0.7402 & 0.6870 & 0.6967 & 0.6311 \\
MiMo-7B-RL       & \textbf{0.8030} & 0.8015 & 0.7919 & 0.7754 & 0.7733 & 0.7448 \\
MiniCPM-4.1-8B   & 0.7702 & 0.7662 & \textbf{0.7715} & 0.7223 & 0.7086 & 0.6862 \\
Nemo-Nano-9B     & 0.7567 & 0.7490 & \textbf{0.7580} & 0.7013 & 0.6922 & 0.6814 \\
OpenThinker-3-7B & 0.7509 & 0.7569 & \textbf{0.7599} & 0.7328 & 0.7165 & 0.6990 \\
Qwen2.5-Coder-7B & \textbf{0.7531} & 0.7460 & 0.7430 & 0.6990 & 0.6987 & 0.7029 \\
Qwen3-8B         & 0.7469 & 0.7400 & \textbf{0.7474} & 0.6966 & 0.6825 & 0.6832 \\
cogito-8B        & \textbf{0.7607} & 0.7579 & 0.7534 & 0.6977 & 0.7000 & 0.7307 \\
gemma2-9b        & 0.7610 & \textbf{0.7654} & 0.7615 & 0.7200 & 0.7187 & 0.7316 \\
glm-4-9b         & \textbf{0.7787} & 0.7742 & 0.7722 & 0.7338 & 0.7347 & 0.7355 \\
granite-3.3-8b   & 0.7599 & 0.7569 & \textbf{0.7615} & 0.7183 & 0.7220 & 0.7225 \\
intlm3-8b        & 0.7572 & \textbf{0.7591} & 0.7470 & 0.7067 & 0.7147 & 0.7237 \\
\midrule
\textbf{Mean}    & \textbf{0.7619} & 0.7598 & 0.7595 & 0.7154 & 0.7130 & 0.7062 \\
\bottomrule
\end{tabular}}
\end{center}
\caption{Small pool --- top 3 and bottom 3 experiments by mean per-model AUC\@.
\textbf{Bold} = best experiment per model among top-3.}
\label{tab:small_auc}
\end{table}

\begin{table}[t]
\begin{center}
\adjustbox{max width=\linewidth}{%
\begin{tabular}{lccccccc}
\toprule
\textbf{Model}
  & \makecell{\#1 Unified\\ Multitask [deberta]}
  & \makecell{\#2 Unified\\ Multitask [qwen-0.6]}
  & \makecell{\#3 Unified\\ Multitask [llama-nemo-v2]}
  & \makecell{last kNN\\{} [mpnet]}
  & \makecell{last-1 kNN\\{} [deberta]}
  & \makecell{last-2 Finetuned\\ deberta v3} \\
\midrule
DeepHermes-3-8B  & 0.2080 & 0.2071 & \textbf{0.2041} & 0.2203 & 0.2194 & 0.2160 \\
DS-R1-Qwen3-8B   & \textbf{0.1712} & 0.1742 & 0.1714 & 0.1859 & 0.1874 & 0.1911 \\
DS-R1D-Qwen-7B   & 0.2036 & \textbf{0.1996} & 0.2013 & 0.2161 & 0.2191 & 0.2288 \\
Fin-R1           & \textbf{0.1838} & 0.1863 & \textbf{0.1838} & 0.2032 & 0.2004 & 0.1993 \\
GLM-Z1-9B        & 0.1775 & 0.1761 & \textbf{0.1743} & 0.1883 & 0.1891 & 0.1938 \\
Intern-S1-mini   & 0.1884 & \textbf{0.1878} & 0.1906 & 0.2009 & 0.2093 & 0.2095 \\
Llama-8B-Inst    & \textbf{0.1964} & 0.1996 & 0.2013 & 0.2154 & 0.2161 & 0.2127 \\
Llama-8B-Med     & \textbf{0.1954} & 0.1966 & 0.2011 & 0.2113 & 0.2077 & 0.2156 \\
Llama-Nemo-8B    & \textbf{0.2009} & 0.2060 & 0.2039 & 0.2195 & 0.2170 & 0.2330 \\
MiMo-7B-RL       & 0.1714 & \textbf{0.1711} & 0.1759 & 0.1796 & 0.1814 & 0.1902 \\
MiniCPM-4.1-8B   & 0.1816 & 0.1810 & \textbf{0.1787} & 0.1950 & 0.1984 & 0.2070 \\
Nemo-Nano-9B     & 0.1773 & 0.1791 & \textbf{0.1752} & 0.1887 & 0.1913 & 0.1951 \\
OpenThinker-3-7B & 0.2001 & \textbf{0.1975} & 0.1983 & 0.2051 & 0.2108 & 0.2174 \\
Qwen2.5-Coder-7B & \textbf{0.2011} & 0.2035 & 0.2049 & 0.2176 & 0.2189 & 0.2172 \\
Qwen3-8B         & 0.1721 & 0.1739 & \textbf{0.1711} & 0.1824 & 0.1860 & 0.1878 \\
cogito-8B        & \textbf{0.1954} & 0.1974 & 0.1988 & 0.2165 & 0.2160 & 0.2090 \\
gemma2-9b        & 0.1967 & 0.1970 & \textbf{0.1966} & 0.2099 & 0.2116 & 0.2084 \\
glm-4-9b         & \textbf{0.1894} & 0.1926 & 0.1931 & 0.2071 & 0.2069 & 0.2066 \\
granite-3.3-8b   & 0.1995 & 0.2005 & \textbf{0.1991} & 0.2133 & 0.2134 & 0.2147 \\
intlm3-8b        & 0.1976 & \textbf{0.1969} & 0.2007 & 0.2132 & 0.2123 & 0.2094 \\
\midrule
\textbf{Mean}    & \textbf{0.1904} & 0.1912 & 0.1912 & 0.2045 & 0.2056 & 0.2081 \\
\bottomrule
\end{tabular}}
\end{center}
\caption{Small pool --- top 3 and bottom 3 experiments by mean per-model Brier score (lower is better).
\textbf{Bold} = best (lowest) experiment per model among top-3.}
\label{tab:small_brier}
\end{table}

\FloatBarrier
\clearpage
\subsection{Mixed Pool}

Tables~\ref{tab:mixed_auc} and~\ref{tab:mixed_brier} show the top-3 and bottom-3 experiments by AUC and Brier score for the mixed pool.

\begin{table}[t]
\begin{center}
\adjustbox{max width=\linewidth}{%
\begin{tabular}{lcccccccccc}
\toprule
\textbf{Experiment} & \textbf{Mean} & opus & gpt5-2 & gpt5-4
  & oss-120b & oss-20b & nano & super & q-122b & q-35b \\
\midrule
\#1 Unified Multitask [llama-nemo-v2]
  & 0.8069 & 0.8226 & 0.7863 & 0.7808 & 0.7912
  & 0.8070 & 0.7990 & 0.8108 & 0.8258 & 0.8390 \\
\#2 Unified Multitask [qwen-0.6]
  & 0.8046 & 0.8127 & 0.7789 & 0.7803 & 0.7832
  & 0.8094 & 0.8059 & 0.8126 & 0.8231 & 0.8353 \\
\#3 Per-Model Multitask [llama-nemo-v2]
  & 0.8023 & 0.8115 & 0.7769 & 0.7726 & 0.7804
  & 0.8118 & 0.8017 & 0.8046 & 0.8222 & 0.8386 \\
\midrule
\#(last-2) Per-Model Multitask + kNN features [mpnet]
  & 0.7440 & 0.7445 & 0.7321 & 0.7032 & 0.7263
  & 0.7530 & 0.7505 & 0.7522 & 0.7643 & 0.7701 \\
\#(last-1) Per-Model Multitask@$\lambda$=0.1 + kNN features [mpnet]
  & 0.7453 & 0.7462 & 0.7291 & 0.7057 & 0.7285
  & 0.7561 & 0.7525 & 0.7493 & 0.7679 & 0.7719 \\
\#(last) kNN [mpnet]
  & 0.7261 & 0.7363 & 0.7070 & 0.6871 & 0.7119
  & 0.7224 & 0.7153 & 0.7364 & 0.7544 & 0.7635 \\
\bottomrule
\end{tabular}}
\end{center}
\caption{Mixed pool --- top 3 and bottom 3 by mean per-model AUC.}
\label{tab:mixed_auc}
\end{table}

\begin{table}[t]
\begin{center}
\adjustbox{max width=\linewidth}{%
\begin{tabular}{lcccccccccc}
\toprule
\textbf{Experiment} & \textbf{Mean} & opus & gpt5-2 & gpt5-4
  & oss-120b & oss-20b & nano & super & q-122b & q-35b \\
\midrule
\#1 Unified Multitask [llama-nemo-v2]
  & 0.1354 & 0.1028 & 0.1337 & 0.1257 & 0.1559
  & 0.1591 & 0.1553 & 0.1455 & 0.1202 & 0.1200 \\
\#2 Unified Multitask [qwen-0.6]
  & 0.1380 & 0.1071 & 0.1377 & 0.1285 & 0.1597
  & 0.1601 & 0.1554 & 0.1453 & 0.1246 & 0.1235 \\
\#3 Per-Model Multitask [llama-nemo-v2]
  & 0.2155 & 0.2389 & 0.2345 & 0.2440 & 0.2107
  & 0.1895 & 0.1997 & 0.2033 & 0.2176 & 0.2013 \\
\midrule
\#(last-2) Per-Model Multitask + kNN features [mpnet]
  & 0.2354 & 0.2603 & 0.2458 & 0.2610 & 0.2261
  & 0.2135 & 0.2197 & 0.2235 & 0.2374 & 0.2311 \\
\#(last-1) Per-Model Multitask@$\lambda$=0.1 + kNN features [mpnet]
  & 0.1605 & 0.1254 & 0.1521 & 0.1472 & 0.1847
  & 0.1866 & 0.1807 & 0.1716 & 0.1460 & 0.1500 \\
\#(last) kNN [mpnet]
  & 0.1657 & 0.1266 & 0.1559 & 0.1469 & 0.1914
  & 0.1969 & 0.1912 & 0.1779 & 0.1491 & 0.1554 \\
\bottomrule
\end{tabular}}
\end{center}
\caption{Mixed pool --- top 3 and bottom 3 by mean per-model Brier score (lower is better).}
\label{tab:mixed_brier}
\end{table}

% ----------------------------------------------------------------
\clearpage
\section{Semantic Baselines Per-Model AUC/Brier Scores}
\label{sec:appendix_permodel}

\subsection{Frontier Pool}

Tables~\ref{tab:frontier_permodel_auc} and~\ref{tab:frontier_permodel_brier} report per-model AUC and Brier scores across semantic baselines for the frontier pool.

\begin{table}[t]
\begin{center}
\adjustbox{max width=\linewidth}{%
\begin{tabular}{lcccccc}
\toprule
\textbf{Model}
  & \makecell{Unified\\ Multitask}
  & \makecell{Matrix\\ Factorization}
  & \makecell{Per-Model\\ Multitask}
  & \makecell{Graph-\\ Router}
  & \makecell{kNN} \\
\midrule
claude-sonnet-4    & 0.8653 & 0.8527 & 0.8633 & 0.8464 & 0.8459 \\
deepseek-r1        & 0.8385 & 0.8331 & 0.8363 & 0.8178 & 0.8275 \\
deepseek-v3        & 0.8049 & 0.8024 & 0.8024 & 0.7782 & 0.7883 \\
gemini-2.5-flash   & 0.8137 & 0.8036 & 0.8099 & 0.8050 & 0.8028 \\
gemini-2.5-pro     & 0.7585 & 0.7439 & 0.7522 & 0.7248 & 0.7437 \\
glm-4.6            & 0.7940 & 0.7866 & 0.7880 & 0.7767 & 0.7789 \\
gpt-5              & 0.7917 & 0.7848 & 0.7875 & 0.7784 & 0.7753 \\
gpt-5-chat         & 0.7804 & 0.7634 & 0.7791 & 0.7605 & 0.7695 \\
kimi-k2            & 0.7991 & 0.7935 & 0.7924 & 0.7785 & 0.7945 \\
openrouter         & 0.8263 & 0.8132 & 0.8196 & 0.8231 & 0.7917 \\
qwen3-235b         & 0.7714 & 0.7598 & 0.7703 & 0.7646 & 0.7583 \\
\midrule
\textbf{Mean}      & \textbf{0.8040} & 0.7943 & 0.8001 & 0.7867 & 0.7888 \\
\bottomrule
\end{tabular}}
\end{center}
\caption{Frontier pool --- per-model AUC across semantic baselines.}
\label{tab:frontier_permodel_auc}
\end{table}

\begin{table}[t]
\begin{center}
\adjustbox{max width=\linewidth}{%
\begin{tabular}{lcccccc}
\toprule
\textbf{Model}
  & \makecell{Unified\\ Multitask}
  & \makecell{Matrix\\ Factorization}
  & \makecell{Per-Model\\ Multitask}
  & \makecell{Graph-\\ Router}
  & \makecell{kNN} \\
\midrule
claude-sonnet-4    & 0.1459 & 0.1504 & 0.1548 & 0.1587 & 0.1536 \\
deepseek-r1        & 0.1593 & 0.1597 & 0.1661 & 0.1744 & 0.1630 \\
deepseek-v3        & 0.1803 & 0.1814 & 0.1826 & 0.1936 & 0.1864 \\
gemini-2.5-flash   & 0.1758 & 0.1790 & 0.1788 & 0.1790 & 0.1794 \\
gemini-2.5-pro     & 0.1855 & 0.1902 & 0.2205 & 0.1977 & 0.1901 \\
glm-4.6            & 0.1830 & 0.1826 & 0.1899 & 0.1889 & 0.1857 \\
gpt-5              & 0.1746 & 0.1756 & 0.2022 & 0.1843 & 0.1797 \\
gpt-5-chat         & 0.1919 & 0.1966 & 0.1944 & 0.1989 & 0.1944 \\
kimi-k2            & 0.1838 & 0.1840 & 0.1867 & 0.1898 & 0.1836 \\
openrouter         & 0.1640 & 0.1696 & 0.1785 & 0.1666 & 0.1808 \\
qwen3-235b         & 0.1873 & 0.1901 & 0.2068 & 0.1876 & 0.1916 \\
\midrule
\textbf{Mean}      & \textbf{0.1756} & 0.1781 & 0.1874 & 0.1836 & 0.1808 \\
\bottomrule
\end{tabular}}
\end{center}
\caption{Frontier pool --- per-model Brier score across semantic baselines (lower is better).}
\label{tab:frontier_permodel_brier}
\end{table}

\FloatBarrier
\clearpage
\subsection{Small Pool}

Tables~\ref{tab:small_permodel_auc} and~\ref{tab:small_permodel_brier} report per-model AUC and Brier scores across semantic baselines for the small pool.

\begin{table}[t]
\begin{center}
\adjustbox{max width=\linewidth}{%
\begin{tabular}{lcccccc}
\toprule
\textbf{Model}
  & \makecell{Unified\\ Multitask}
  & \makecell{Matrix\\ Factorization}
  & \makecell{Per-Model\\ Multitask}
  & \makecell{Graph-\\ Router}
  & \makecell{kNN} \\
\midrule
DeepHermes-3-8B    & 0.7390 & 0.7399 & 0.7335 & 0.7055 & 0.7054 \\
DS-R1-Qwen3-8B     & 0.7709 & 0.7503 & 0.7688 & 0.7537 & 0.7436 \\
DS-R1D-Qwen-7B     & 0.7532 & 0.7405 & 0.7482 & 0.7316 & 0.7244 \\
Fin-R1             & 0.7932 & 0.7763 & 0.7885 & 0.7698 & 0.7625 \\
GLM-Z1-9B          & 0.7698 & 0.7545 & 0.7615 & 0.7616 & 0.7399 \\
Intern-S1-mini     & 0.7482 & 0.7455 & 0.7554 & 0.7415 & 0.7218 \\
Llama-8B-Inst      & 0.7552 & 0.7419 & 0.7511 & 0.7442 & 0.7267 \\
Llama-8B-Med       & 0.7538 & 0.7513 & 0.7517 & 0.7223 & 0.7396 \\
Llama-Nemo-8B      & 0.7402 & 0.7310 & 0.7402 & 0.7244 & 0.7196 \\
MiMo-7B-RL         & 0.7919 & 0.7891 & 0.7949 & 0.7874 & 0.7945 \\
MiniCPM-4.1-8B     & 0.7715 & 0.7534 & 0.7664 & 0.7539 & 0.7328 \\
Nemo-Nano-9B       & 0.7580 & 0.7432 & 0.7446 & 0.7418 & 0.7162 \\
OpenThinker-3-7B   & 0.7599 & 0.7433 & 0.7529 & 0.7376 & 0.7397 \\
Qwen2.5-Coder-7B   & 0.7430 & 0.7417 & 0.7406 & 0.7246 & 0.7068 \\
Qwen3-8B           & 0.7474 & 0.7188 & 0.7334 & 0.7208 & 0.7107 \\
cogito-8B          & 0.7534 & 0.7455 & 0.7566 & 0.7462 & 0.7275 \\
gemma2-9b          & 0.7615 & 0.7513 & 0.7532 & 0.7401 & 0.7246 \\
glm-4-9b           & 0.7722 & 0.7628 & 0.7719 & 0.7547 & 0.7547 \\
granite-3.3-8b     & 0.7615 & 0.7471 & 0.7498 & 0.7292 & 0.7386 \\
intlm3-8b          & 0.7470 & 0.7430 & 0.7432 & 0.7256 & 0.7273 \\
\midrule
\textbf{Mean}      & \textbf{0.7595} & 0.7485 & 0.7553 & 0.7408 & 0.7328 \\
\bottomrule
\end{tabular}}
\end{center}
\caption{Small pool --- per-model AUC across semantic baselines.}
\label{tab:small_permodel_auc}
\end{table}

\begin{table}[t]
\begin{center}
\adjustbox{max width=\linewidth}{%
\begin{tabular}{lcccccc}
\toprule
\textbf{Model}
  & \makecell{Unified\\ Multitask}
  & \makecell{Matrix\\ Factorization}
  & \makecell{Per-Model\\ Multitask}
  & \makecell{Graph-\\ Router}
  & \makecell{kNN} \\
\midrule
DeepHermes-3-8B    & 0.2041 & 0.2056 & 0.2077 & 0.2167 & 0.2153 \\
DS-R1-Qwen3-8B     & 0.1714 & 0.1784 & 0.1729 & 0.1823 & 0.1793 \\
DS-R1D-Qwen-7B     & 0.2013 & 0.2061 & 0.2031 & 0.2092 & 0.2109 \\
Fin-R1             & 0.1838 & 0.1893 & 0.1851 & 0.1926 & 0.1940 \\
GLM-Z1-9B          & 0.1743 & 0.1791 & 0.1772 & 0.1826 & 0.1838 \\
Intern-S1-mini     & 0.1906 & 0.1923 & 0.1893 & 0.1947 & 0.1989 \\
Llama-8B-Inst      & 0.2013 & 0.2051 & 0.2019 & 0.2065 & 0.2098 \\
Llama-8B-Med       & 0.2011 & 0.2005 & 0.2003 & 0.2100 & 0.2041 \\
Llama-Nemo-8B      & 0.2039 & 0.2073 & 0.2041 & 0.2084 & 0.2100 \\
MiMo-7B-RL         & 0.1759 & 0.1756 & 0.1764 & 0.1806 & 0.1732 \\
MiniCPM-4.1-8B     & 0.1787 & 0.1858 & 0.1818 & 0.1879 & 0.1917 \\
Nemo-Nano-9B       & 0.1752 & 0.1797 & 0.1789 & 0.1819 & 0.1856 \\
OpenThinker-3-7B   & 0.1983 & 0.2029 & 0.1992 & 0.2051 & 0.2030 \\
Qwen2.5-Coder-7B   & 0.2049 & 0.2053 & 0.2038 & 0.2106 & 0.2151 \\
Qwen3-8B           & 0.1711 & 0.1779 & 0.1742 & 0.1841 & 0.1799 \\
cogito-8B          & 0.1988 & 0.2016 & 0.1975 & 0.2023 & 0.2066 \\
gemma2-9b          & 0.1966 & 0.1994 & 0.1986 & 0.2035 & 0.2074 \\
glm-4-9b           & 0.1931 & 0.1966 & 0.1928 & 0.2020 & 0.1995 \\
granite-3.3-8b     & 0.1991 & 0.2049 & 0.2041 & 0.2121 & 0.2071 \\
intlm3-8b          & 0.2007 & 0.2027 & 0.2012 & 0.2085 & 0.2061 \\
\midrule
\textbf{Mean}      & \textbf{0.1912} & 0.1948 & 0.1925 & 0.1991 & 0.1991 \\
\bottomrule
\end{tabular}}
\end{center}
\caption{Small pool --- per-model Brier score across semantic baselines (lower is better).}
\label{tab:small_permodel_brier}
\end{table}

\FloatBarrier
\clearpage
\subsection{Mixed Pool}

Tables~\ref{tab:mixed_permodel_auc} and~\ref{tab:mixed_permodel_brier} report per-model AUC and Brier scores across semantic baselines for the mixed pool.

\begin{table}[t]
\begin{center}
\adjustbox{max width=\linewidth}{%
\begin{tabular}{lcccccc}
\toprule
\textbf{Model}
  & \makecell{Unified\\ Multitask}
  & \makecell{Matrix\\ Factorization}
  & \makecell{Per-Model\\ Multitask}
  & \makecell{Graph-\\ Router}
  & \makecell{kNN} \\
\midrule
claude-opus-4      & 0.8226 & 0.7847 & 0.8115 & 0.7995 & 0.7901 \\
gpt-5-2            & 0.7863 & 0.7578 & 0.7769 & 0.7732 & 0.7521 \\
gpt-5-4            & 0.7808 & 0.7591 & 0.7726 & 0.7627 & 0.7422 \\
gpt-oss-120b       & 0.7912 & 0.7633 & 0.7804 & 0.7743 & 0.7425 \\
gpt-oss-20b        & 0.8070 & 0.7912 & 0.8118 & 0.8067 & 0.7757 \\
nemotron-nano-30b  & 0.7990 & 0.7802 & 0.8017 & 0.7924 & 0.7654 \\
nemotron-super-120b & 0.8108 & 0.7868 & 0.8046 & 0.7965 & 0.7799 \\
qwen-3.5-122b      & 0.8258 & 0.7975 & 0.8222 & 0.8094 & 0.8024 \\
qwen-3.5-35b       & 0.8390 & 0.8126 & 0.8386 & 0.8277 & 0.8061 \\
\midrule
\textbf{Mean}      & \textbf{0.8069} & 0.7815 & 0.8023 & 0.7936 & 0.7729 \\
\bottomrule
\end{tabular}}
\end{center}
\caption{Mixed pool --- per-model AUC across semantic baselines.}
\label{tab:mixed_permodel_auc}
\end{table}

\begin{table}[t]
\begin{center}
\adjustbox{max width=\linewidth}{%
\begin{tabular}{lcccccc}
\toprule
\textbf{Model}
  & \makecell{Unified\\ Multitask}
  & \makecell{Matrix\\ Factorization}
  & \makecell{Per-Model\\ Multitask}
  & \makecell{Graph-\\ Router}
  & \makecell{kNN} \\
\midrule
claude-opus-4      & 0.1028 & 0.1079 & 0.2389 & 0.1165 & 0.1143 \\
gpt-5-2            & 0.1337 & 0.1370 & 0.2345 & 0.1473 & 0.1446 \\
gpt-5-4            & 0.1257 & 0.1291 & 0.2440 & 0.1394 & 0.1369 \\
gpt-oss-120b       & 0.1559 & 0.1644 & 0.2107 & 0.1637 & 0.1791 \\
gpt-oss-20b        & 0.1591 & 0.1658 & 0.1895 & 0.1651 & 0.1793 \\
nemotron-nano-30b  & 0.1553 & 0.1619 & 0.1997 & 0.1627 & 0.1736 \\
nemotron-super-120b & 0.1455 & 0.1514 & 0.2033 & 0.1537 & 0.1629 \\
qwen-3.5-122b      & 0.1202 & 0.1249 & 0.2176 & 0.1296 & 0.1353 \\
qwen-3.5-35b       & 0.1200 & 0.1273 & 0.2013 & 0.1298 & 0.1421 \\
\midrule
\textbf{Mean}      & \textbf{0.1354} & 0.1411 & 0.2155 & 0.1453 & 0.1520 \\
\bottomrule
\end{tabular}}
\end{center}
\caption{Mixed pool --- per-model Brier score across semantic baselines (lower is better).}
\label{tab:mixed_permodel_brier}
\end{table}

% ----------------------------------------------------------------
\clearpage
\section{Seed Variability and Bootstrap Confidence Intervals}
\label{sec:appendix_uncertainty}

We quantify two complementary sources of uncertainty. First, we train each
stochastic router with 10 random seeds and report the sample standard deviation
across runs. Second, holding the originally submitted prediction files fixed,
we perform a query-level nonparametric bootstrap with $N{=}1000$ resamples and
report percentile 95\% confidence intervals. Seed variation measures training
instability, whereas the bootstrap measures test-sample uncertainty; because
the seed study uses newly trained single-seed models and the bootstrap uses the
submitted prediction snapshot (a five-network ensemble for SharedTrunkNet),
their point estimates are not expected to be identical.

\subsection{Variation Across Training Seeds}

Table~\ref{tab:seed_variability_all} reports mean $\pm$ sample standard
deviation across 10 independently trained seeds. kNN voting is deterministic
after feature extraction and therefore has no seed deviation.

\begin{table}[H]
\begin{center}
\scriptsize
\setlength{\tabcolsep}{3.2pt}
\renewcommand{\arraystretch}{1.08}
\adjustbox{max width=\textwidth}{%
\begin{tabular}{llcccc}
\toprule
\textbf{Pool} & \textbf{System} & \textbf{AUC} & \textbf{Brier}
  & \makecell{\textbf{Max-$\lambda$}\\\textbf{accuracy}} & \textbf{P-AUCCC} \\
\midrule
\multirow{6}{*}{Frontier}
 & SharedTrunkNet (Fisher) & $0.8523{\pm}0.0014$ & $0.1532{\pm}0.0007$ & $0.7559{\pm}0.0046$ & $0.4308{\pm}0.0067$ \\
 & Per-Model Multitask     & $0.7982{\pm}0.0012$ & $0.1883{\pm}0.0006$ & $0.6726{\pm}0.0123$ & $0.3678{\pm}0.0062$ \\
 & Unified Multitask       & $0.8005{\pm}0.0020$ & $0.1767{\pm}0.0010$ & $0.7326{\pm}0.0053$ & $0.4150{\pm}0.0057$ \\
 & Matrix Factorization    & $0.7932{\pm}0.0007$ & $0.1847{\pm}0.0004$ & $0.7108{\pm}0.0066$ & $0.3918{\pm}0.0044$ \\
 & kNN voting ($k{=}25$)   & $0.7888$             & $0.1808$             & $0.7120$             & $0.4072$ \\
 & GraphRouter             & $0.7860{\pm}0.0018$ & $0.1841{\pm}0.0010$ & $0.7037{\pm}0.0051$ & $0.3889{\pm}0.0047$ \\
\midrule
\multirow{6}{*}{Small}
 & SharedTrunkNet (Fisher) & $0.8215{\pm}0.0009$ & $0.1664{\pm}0.0004$ & $0.7386{\pm}0.0043$ & $0.5357{\pm}0.0040$ \\
 & Per-Model Multitask     & $0.7550{\pm}0.0013$ & $0.1927{\pm}0.0005$ & $0.7309{\pm}0.0037$ & $0.5061{\pm}0.0096$ \\
 & Unified Multitask       & $0.7585{\pm}0.0021$ & $0.1912{\pm}0.0008$ & $0.7371{\pm}0.0047$ & $0.5276{\pm}0.0076$ \\
 & Matrix Factorization    & $0.7476{\pm}0.0004$ & $0.2063{\pm}0.0001$ & $0.6689{\pm}0.0051$ & $0.4514{\pm}0.0054$ \\
 & kNN voting ($k{=}25$)   & $0.7328$             & $0.1991$             & $0.7370$             & $0.5122$ \\
 & GraphRouter             & $0.7397{\pm}0.0014$ & $0.1992{\pm}0.0005$ & $0.7090{\pm}0.0049$ & $0.4635{\pm}0.0072$ \\
\midrule
\multirow{6}{*}{Mixed}
 & SharedTrunkNet (Fisher) & $0.8780{\pm}0.0015$ & $0.1132{\pm}0.0007$ & $0.8230{\pm}0.0039$ & $0.2143{\pm}0.0098$ \\
 & Per-Model Multitask     & $0.8013{\pm}0.0008$ & $0.2147{\pm}0.0004$ & $0.7530{\pm}0.0058$ & $0.1629{\pm}0.0073$ \\
 & Unified Multitask       & $0.8050{\pm}0.0022$ & $0.1358{\pm}0.0005$ & $0.8256{\pm}0.0041$ & $0.1932{\pm}0.0116$ \\
 & Matrix Factorization    & $0.7807{\pm}0.0021$ & $0.1947{\pm}0.0010$ & $0.7780{\pm}0.0037$ & $0.1720{\pm}0.0130$ \\
 & kNN voting ($k{=}25$)   & $0.7729$             & $0.1520$             & $0.8023$             & $0.1874$ \\
 & GraphRouter             & $0.7925{\pm}0.0015$ & $0.1468{\pm}0.0004$ & $0.7910{\pm}0.0056$ & $0.1798{\pm}0.0077$ \\
\bottomrule
\end{tabular}}
\end{center}
\caption{Training-seed variability for the six headline routers. Semantic
baselines use the calibrated configurations reported in the main results.
Values are mean $\pm$ sample standard deviation over 10 seeds, except the
deterministic kNN baseline.}
\label{tab:seed_variability_all}
\end{table}

For SharedTrunkNet, the corresponding mean [minimum, maximum] ranges are:
frontier AUC $0.8523[0.8497,0.8545]$, Brier
$0.1532[0.1521,0.1546]$, max-$\lambda$ accuracy
$0.7559[0.7488,0.7645]$, and P-AUCCC $0.4308[0.4162,0.4405]$; small
$0.8215[0.8202,0.8230]$, $0.1664[0.1656,0.1670]$,
$0.7386[0.7323,0.7459]$, and $0.5357[0.5307,0.5415]$; and mixed
$0.8780[0.8751,0.8804]$, $0.1132[0.1123,0.1145]$,
$0.8230[0.8161,0.8290]$, and $0.2143[0.2053,0.2344]$, respectively.
These brackets are seed extremes, not confidence intervals.

\subsection{Query-Level Bootstrap}

Tables~\ref{tab:bootstrap_frontier}, \ref{tab:bootstrap_small}, and
\ref{tab:bootstrap_mixed} give the query-level bootstrap results for the
frontier, small, and mixed pools, respectively. Each resample draws held-out queries with
replacement and recomputes macro AUC, macro Brier score, the maximum accuracy
along the $\lambda$ sweep, and P-AUCCC under the cost model in
Equation~\ref{eq:cost_est}.

\begin{table}[H]
\begin{center}
\scriptsize
\setlength{\tabcolsep}{2.5pt}
\adjustbox{max width=\textwidth}{%
\begin{tabular}{lcccc}
\toprule
\textbf{System} & \textbf{AUC [95\% CI]} & \textbf{Brier [95\% CI]}
 & \makecell{\textbf{Max-$\lambda$ accuracy}\\\textbf{[95\% CI]}}
 & \textbf{P-AUCCC [95\% CI]} \\
\midrule
SharedTrunkNet (Fisher) & $0.8560[0.8437,0.8674]$ & $0.1509[0.1446,0.1578]$ & $0.7645[0.7454,0.7876]$ & $0.4536[0.4130,0.4953]$ \\
Per-Model Multitask     & $0.8001[0.7861,0.8157]$ & $0.1874[0.1799,0.1941]$ & $0.6801[0.6562,0.7039]$ & $0.3737[0.3328,0.4134]$ \\
Unified Multitask       & $0.8040[0.7883,0.8190]$ & $0.1756[0.1681,0.1833]$ & $0.7345[0.7154,0.7590]$ & $0.4266[0.3830,0.4688]$ \\
Matrix Factorization    & $0.7943[0.7784,0.8103]$ & $0.1844[0.1769,0.1913]$ & $0.7052[0.6828,0.7291]$ & $0.3895[0.3483,0.4338]$ \\
kNN voting ($k{=}25$)   & $0.7888[0.7739,0.8037]$ & $0.1808[0.1747,0.1870]$ & $0.7120[0.6909,0.7345]$ & $0.4072[0.3626,0.4463]$ \\
GraphRouter             & $0.7867[0.7714,0.8013]$ & $0.1836[0.1779,0.1898]$ & $0.6984[0.6780,0.7223]$ & $0.3910[0.3471,0.4321]$ \\
\bottomrule
\end{tabular}}
\end{center}
\caption{Frontier-pool query-level bootstrap intervals ($N{=}1000$,
percentile method).}
\label{tab:bootstrap_frontier}
\end{table}

\begin{table}[H]
\begin{center}
\scriptsize
\setlength{\tabcolsep}{2.5pt}
\adjustbox{max width=\textwidth}{%
\begin{tabular}{lcccc}
\toprule
\textbf{System} & \textbf{AUC [95\% CI]} & \textbf{Brier [95\% CI]}
 & \makecell{\textbf{Max-$\lambda$ accuracy}\\\textbf{[95\% CI]}}
 & \textbf{P-AUCCC [95\% CI]} \\
\midrule
SharedTrunkNet (Fisher) & $0.8260[0.8165,0.8349]$ & $0.1642[0.1598,0.1685]$ & $0.7548[0.7379,0.7741]$ & $0.5514[0.5276,0.5742]$ \\
Per-Model Multitask     & $0.7553[0.7432,0.7665]$ & $0.1925[0.1886,0.1966]$ & $0.7393[0.7224,0.7590]$ & $0.5240[0.5012,0.5480]$ \\
Unified Multitask       & $0.7595[0.7468,0.7728]$ & $0.1912[0.1864,0.1958]$ & $0.7412[0.7266,0.7609]$ & $0.5308[0.5082,0.5531]$ \\
Matrix Factorization    & $0.7485[0.7367,0.7599]$ & $0.2059[0.2019,0.2099]$ & $0.6689[0.6505,0.6909]$ & $0.4785[0.4515,0.5045]$ \\
kNN voting ($k{=}25$)   & $0.7328[0.7204,0.7436]$ & $0.1991[0.1957,0.2031]$ & $0.7370[0.7201,0.7553]$ & $0.5122[0.4840,0.5351]$ \\
GraphRouter             & $0.7408[0.7300,0.7508]$ & $0.1991[0.1955,0.2026]$ & $0.7060[0.6872,0.7266]$ & $0.4842[0.4592,0.5100]$ \\
\bottomrule
\end{tabular}}
\end{center}
\caption{Small-pool query-level bootstrap intervals ($N{=}1000$, percentile
method).}
\label{tab:bootstrap_small}
\end{table}

\begin{table}[H]
\begin{center}
\scriptsize
\setlength{\tabcolsep}{2.5pt}
\adjustbox{max width=\textwidth}{%
\begin{tabular}{lcccc}
\toprule
\textbf{System} & \textbf{AUC [95\% CI]} & \textbf{Brier [95\% CI]}
 & \makecell{\textbf{Max-$\lambda$ accuracy}\\\textbf{[95\% CI]}}
 & \textbf{P-AUCCC [95\% CI]} \\
\midrule
SharedTrunkNet (Fisher) & $0.8817[0.8682,0.8951]$ & $0.1111[0.1048,0.1183]$ & $0.8336[0.8175,0.8502]$ & $0.2201[0.1914,0.2458]$ \\
Per-Model Multitask     & $0.8023[0.7847,0.8178]$ & $0.2155[0.2082,0.2233]$ & $0.7627[0.7438,0.7802]$ & $0.1290[0.1040,0.1557]$ \\
Unified Multitask       & $0.8069[0.7899,0.8252]$ & $0.1354[0.1281,0.1428]$ & $0.8327[0.8175,0.8470]$ & $0.1808[0.1485,0.2102]$ \\
Matrix Factorization    & $0.7815[0.7636,0.8007]$ & $0.1939[0.1882,0.1996]$ & $0.7770[0.7604,0.7949]$ & $0.1519[0.1173,0.1834]$ \\
kNN voting ($k{=}25$)   & $0.7729[0.7563,0.7889]$ & $0.1520[0.1449,0.1593]$ & $0.8023[0.7853,0.8203]$ & $0.1874[0.1610,0.2125]$ \\
GraphRouter             & $0.7936[0.7769,0.8094]$ & $0.1453[0.1385,0.1529]$ & $0.7931[0.7751,0.8092]$ & $0.1857[0.1574,0.2144]$ \\
\bottomrule
\end{tabular}}
\end{center}
\caption{Mixed-pool query-level bootstrap intervals ($N{=}1000$, percentile
method).}
\label{tab:bootstrap_mixed}
\end{table}

For SharedTrunkNet, the largest one-sided distance from a point estimate to a
bootstrap endpoint across the three pools is 0.0135 for AUC, 0.0072 for Brier,
0.0231 for max-$\lambda$ accuracy, and 0.0417 for P-AUCCC. The paired
bootstrap below directly tests system differences while preserving the
per-query correlation between routers.

\FloatBarrier
\clearpage

% ----------------------------------------------------------------
\section{Paired Bootstrap Significance Against Unified Multitask}
\label{sec:appendix_paired_bootstrap}

Because SharedTrunkNet and Unified Multitask are evaluated on the same held-out
queries, comparing their marginal bootstrap confidence intervals discards the
correlation induced by the shared query sample and is unnecessarily
conservative. We therefore run a paired query-level bootstrap on the metric
difference (SharedTrunkNet/Fisher $-$ Unified Multitask), using identical query
indices for both systems in each of $N{=}1000$ resamples.
Table~\ref{tab:paired_bootstrap_unified} reports the point differences and
percentile 95\% confidence intervals.

\begin{table}[H]
\begin{center}
\small
\setlength{\tabcolsep}{5pt}
\adjustbox{max width=\textwidth}{%
\begin{tabular}{llrrlc}
\toprule
\textbf{Pool} & \textbf{Metric}
  & \makecell{\textbf{SharedTrunkNet}\\\textbf{(Fisher)}}
  & \makecell{\textbf{Unified}\\\textbf{Multitask}}
  & \makecell{\textbf{Difference}\\\textbf{[95\% CI]}}
  & \textbf{Sig.} \\
\midrule
Frontier & AUC (pp)
  & 85.60 & 80.40
  & $\mathbf{+5.20}\;[\mathbf{+4.28},\mathbf{+6.10}]$ & Yes \\
Frontier & Brier ($\times 10^{-2}$)
  & 15.09 & 17.56
  & $\mathbf{-2.47}\;[\mathbf{-2.94},\mathbf{-2.05}]$ & Yes \\
Frontier & Max-$\lambda$ accuracy (pp)
  & 76.45 & 73.45
  & $\mathbf{+3.00}\;[\mathbf{+1.57},\mathbf{+4.29}]$ & Yes \\
Frontier & P-AUCCC (pp)
  & 45.36 & 42.66
  & $\mathbf{+2.69}\;[\mathbf{+1.16},\mathbf{+4.32}]$ & Yes \\
\midrule
Small & AUC (pp)
  & 82.60 & 75.95
  & $\mathbf{+6.65}\;[\mathbf{+5.74},\mathbf{+7.49}]$ & Yes \\
Small & Brier ($\times 10^{-2}$)
  & 16.42 & 19.12
  & $\mathbf{-2.70}\;[\mathbf{-3.05},\mathbf{-2.32}]$ & Yes \\
Small & Max-$\lambda$ accuracy (pp)
  & 75.48 & 74.12
  & $+1.36\;[+0.00,+2.54]$ & No \\
Small & P-AUCCC (pp)
  & 55.14 & 53.08
  & $\mathbf{+2.06}\;[\mathbf{+0.79},\mathbf{+3.32}]$ & Yes \\
\midrule
Mixed & AUC (pp)
  & 88.17 & 80.69
  & $\mathbf{+7.48}\;[\mathbf{+6.16},\mathbf{+8.88}]$ & Yes \\
Mixed & Brier ($\times 10^{-2}$)
  & 11.11 & 13.54
  & $\mathbf{-2.43}\;[\mathbf{-2.86},\mathbf{-2.03}]$ & Yes \\
Mixed & Max-$\lambda$ accuracy (pp)
  & 83.36 & 83.27
  & $+0.09\;[-0.69,+0.83]$ & No \\
Mixed & P-AUCCC (pp)
  & 22.01 & 18.08
  & $\mathbf{+3.94}\;[\mathbf{+1.50},\mathbf{+6.37}]$ & Yes \\
\bottomrule
\end{tabular}}
\end{center}
\caption{Paired query-level bootstrap comparison of SharedTrunkNet
(Fisher-selected) and Unified Multitask using identical query indices in each
of $N{=}1000$ resamples. Differences are SharedTrunkNet minus Unified
Multitask. AUC, max-$\lambda$ accuracy, and P-AUCCC are reported in percentage
points (pp); Brier scores are scaled by $10^{-2}$. Bold differences indicate
that the 95\% confidence interval excludes zero in the favorable direction
(positive for all metrics except Brier).}
\label{tab:paired_bootstrap_unified}
\end{table}

The paired confidence interval excludes zero in the favorable direction for 10
of the 12 pool--metric comparisons. The only non-significant cells are
max-$\lambda$ accuracy on the small and mixed pools. P-AUCCC, which integrates
performance over the cost--accuracy curve rather than using only the
maximum-accuracy operating point, remains significantly higher for
SharedTrunkNet in all three pools.

\FloatBarrier

% ----------------------------------------------------------------
\clearpage
\section{Accuracy Versus Routing Ratio}
\label{sec:appendix_routing_ratio}

To isolate routing quality from cost-aware $\lambda$ tuning, we perform a
forced routing-ratio sweep. For each pool, the strong target is the model with
the highest marginal test accuracy. Queries are ranked by each router's
predicted probability for that target; the top $rN$ queries are assigned to the
strong target, while the remainder are assigned by the router's argmax over
the other targets. Varying $r$ from 0 to 1 therefore measures routed accuracy
at a fixed allocation rate independently of model prices.

\begin{figure}[H]
  \centering
  \begin{minipage}[t]{0.49\linewidth}
    \centering
    \includegraphics[width=\linewidth]{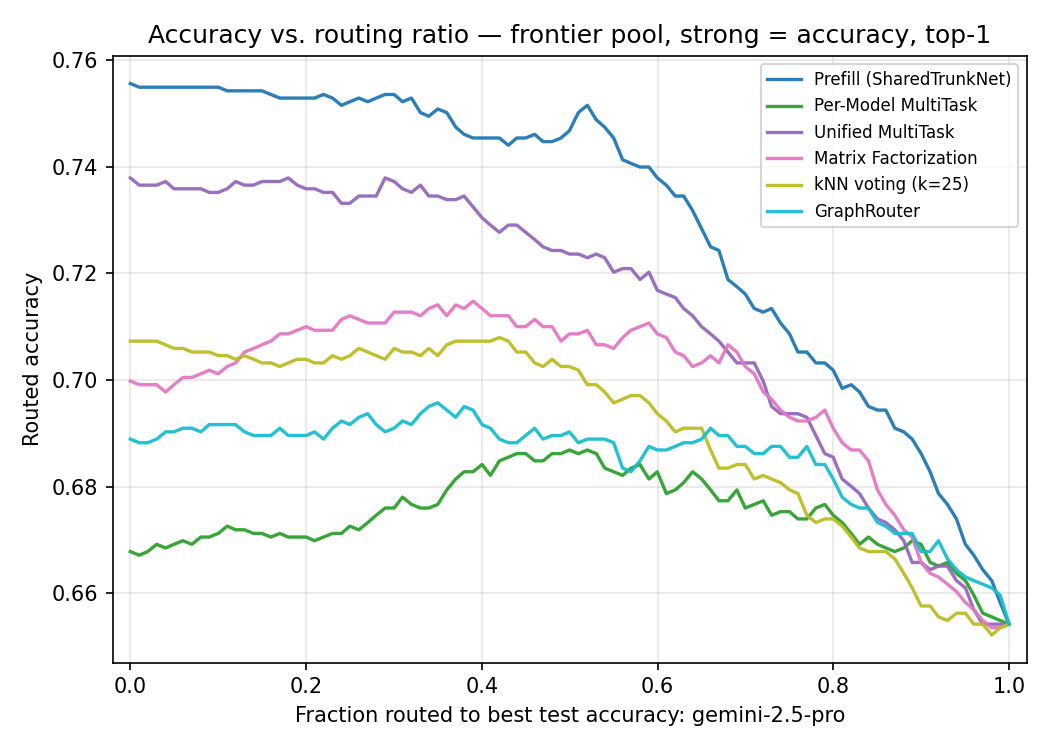}
    \small (a) Frontier pool: Gemini-2.5-Pro is the strong target.
  \end{minipage}
  \hfill
  \begin{minipage}[t]{0.49\linewidth}
    \centering
    \includegraphics[width=\linewidth]{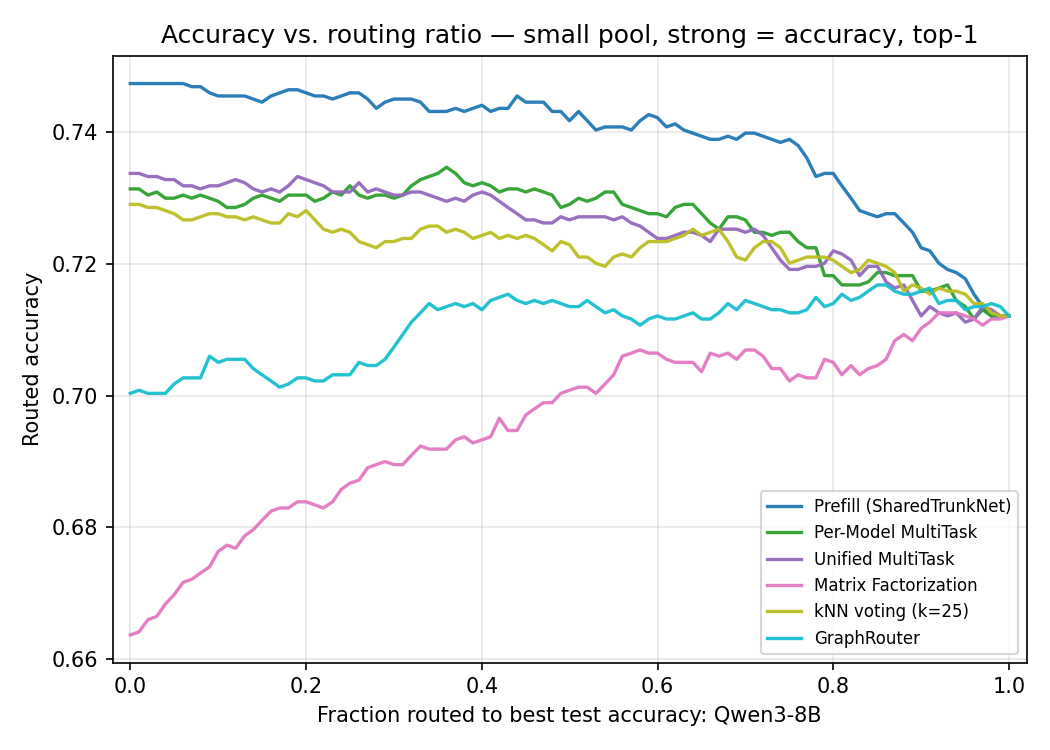}
    \small (b) Small pool: Qwen3-8B is the strong target.
  \end{minipage}

  \vspace{0.8em}
  \begin{minipage}[t]{0.56\linewidth}
    \centering
    \includegraphics[width=\linewidth]{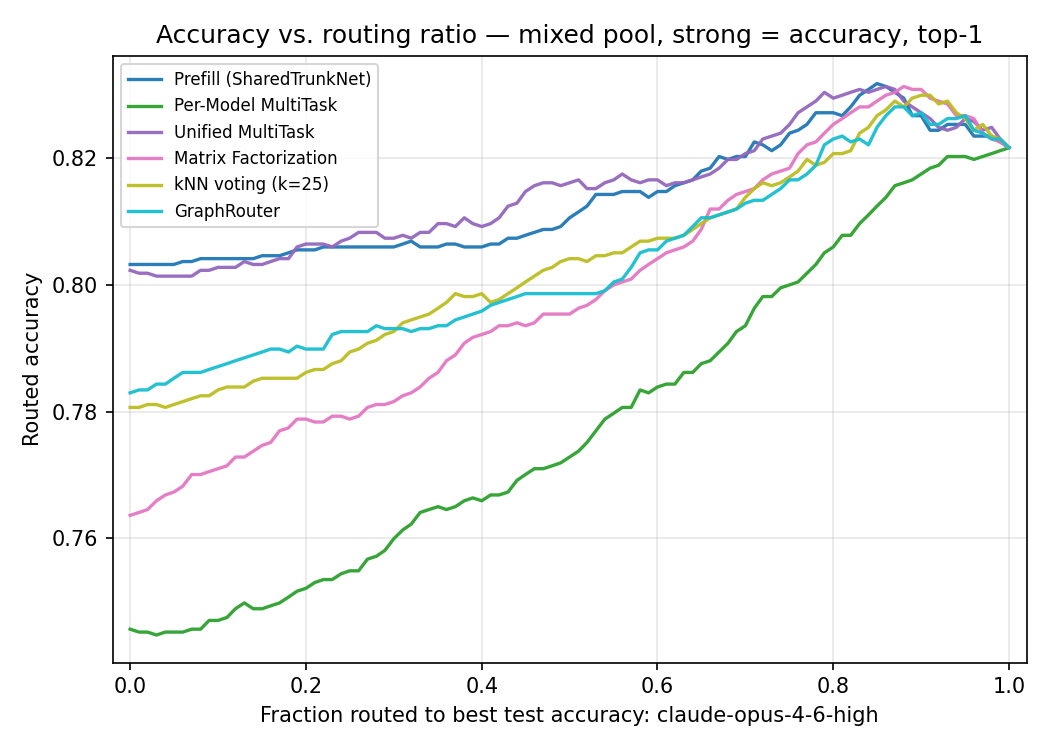}
    \small (c) Mixed pool: Claude Opus 4.6 is the strong target.
  \end{minipage}
  \caption{Routed accuracy as a function of the fraction of queries forced to
  the highest-marginal-accuracy target. The sweep fixes the routing ratio and
  thus removes cost-aware threshold tuning from the comparison. All methods
  converge at $r=1$, where every query is sent to the same strong target.}
  \label{fig:routing_ratio_sweeps}
\end{figure}

As shown in Figure~\ref{fig:routing_ratio_sweeps}, SharedTrunkNet leads the
semantic baselines throughout the practically relevant low-ratio regime on the
frontier and small pools. On the mixed pool,
it closely tracks Unified Multitask, with differences within 0.5 percentage
points at the representative routing ratios. These trends are consistent with
the P-AUCCC comparisons while exposing routing quality without cost-aware
operating-point selection.

\FloatBarrier

% ----------------------------------------------------------------
\clearpage
\section{End-to-End Routing Cost and Latency}
\label{sec:appendix_end_to_end_cost}

The main cost curves price target inference. Here we additionally charge the
prefill router for the encoder's actual input sequence length (ISL), while
pricing the semantic router at zero dollars to give it a strict advantage. On
the 1,469-query frontier test set, Qwen3.5-35B-A3B processes a mean of 1,273.3
input tokens at \$0.1625 per million input tokens. Its mean prefill cost is
\$0.000207 per query; combined with \$0.027115 of routed target inference,
this yields \$0.027322 per query, of which the encoder accounts for 0.76\%.
Table~\ref{tab:end_to_end_cost_points} summarizes these end-to-end costs at two
operating points.

\begin{table}[H]
\begin{center}
\small
\setlength{\tabcolsep}{5pt}
\adjustbox{max width=\textwidth}{%
\begin{tabular}{lrrrrrr}
\toprule
\textbf{Operating point}
 & \makecell{\textbf{Prefill}\\\textbf{total (\$)}}
 & \makecell{\textbf{Semantic}\\\textbf{total (\$)}}
 & \makecell{\textbf{Best single}\\\textbf{total (\$)}}
 & \makecell{\textbf{Savings vs.}\\\textbf{semantic}}
 & \makecell{\textbf{Savings vs.}\\\textbf{best single}}
 & \makecell{\textbf{Accuracy gain}\\\textbf{vs. semantic (pp)}} \\
\midrule
Argmax accuracy & 40.14 & 44.51 & 81.79 & 9.8\% & 50.9\% & $+5.85$ \\
$\lambda=0.5$   & 32.98 & 39.36 & 81.79 & 16.2\% & 59.7\% & $+6.26$ \\
\bottomrule
\end{tabular}}
\end{center}
\caption{End-to-end frontier-pool cost at two operating points. Dollar
figures are totals over all 1,469 test queries and include encoder prefill for
SharedTrunkNet; semantic-router compute is priced at \$0. The best-single
reference is Gemini-2.5-Pro. Accuracy differences are percentage points.}
\label{tab:end_to_end_cost_points}
\end{table}

Table~\ref{tab:end_to_end_latency} separately accounts for user-visible
latency. We use published mean time-to-first-token (TTFT) and throughput,
compute target completion time as target TTFT plus the query's observed output
length divided by throughput, and then add the sequential routing pre-step.
Because routers select different targets, their mean target completion times
also differ; the overhead-share column isolates the fraction due only to the
routing pre-step.

\begin{table}[H]
\begin{center}
\small
\setlength{\tabcolsep}{4.5pt}
\adjustbox{max width=\textwidth}{%
\begin{tabular}{lrrrrr}
\toprule
\textbf{Scenario}
 & \makecell{\textbf{Routing}\\\textbf{latency (ms)}}
 & \makecell{\textbf{Mean target}\\\textbf{completion (s)}}
 & \makecell{\textbf{Routing latency}\\\textbf{share}}
 & \makecell{\textbf{Total cost}\\\textbf{per query (\$)}}
 & \makecell{\textbf{Argmax}\\\textbf{accuracy}} \\
\midrule
Prefill routing & 510 & 64.287 & 0.79\% & 0.027322 & 75.97\% \\
Semantic routing & 159 & 53.686 & 0.30\% & 0.030301 & 70.12\% \\
Best single model & 0 & 61.967 & 0.00\% & 0.055674 & 65.42\% \\
\bottomrule
\end{tabular}}
\end{center}
\caption{Frontier-pool latency and cost at argmax accuracy. Prefill routing
adds 351\,ms relative to the semantic embedding step, but saves
\$0.002979 per query and improves routed accuracy by 5.85 percentage points.
The semantic encoder retains its measured latency but is assigned zero dollar
cost.}
\label{tab:end_to_end_latency}
\end{table}

Encoder prefill adds 0.76\% to the routed dollar cost and 0.79\% to mean
end-to-end wall-clock time at this operating point. In the normalized
P-AUCCC calculation, adding the same mean prefill term to every point
translates the SharedTrunkNet curve horizontally but leaves its normalized
integral unchanged to the reported precision. Thus, charging the encoder does
not alter the ordering in the frontier-pool P-AUCCC comparison.

\FloatBarrier
\clearpage

% ----------------------------------------------------------------
\section{Accuracy--Cost Curves}
\label{sec:appendix_curves}
\subsection{Small Pool}

\begin{figure}[t]
  \centering
  \begin{minipage}[t]{0.58\linewidth}
    \centering
    {\setlength{\fboxsep}{0pt}\setlength{\fboxrule}{0.2pt}%
    \fbox{\includegraphics[width=\dimexpr\linewidth-2\fboxrule\relax]{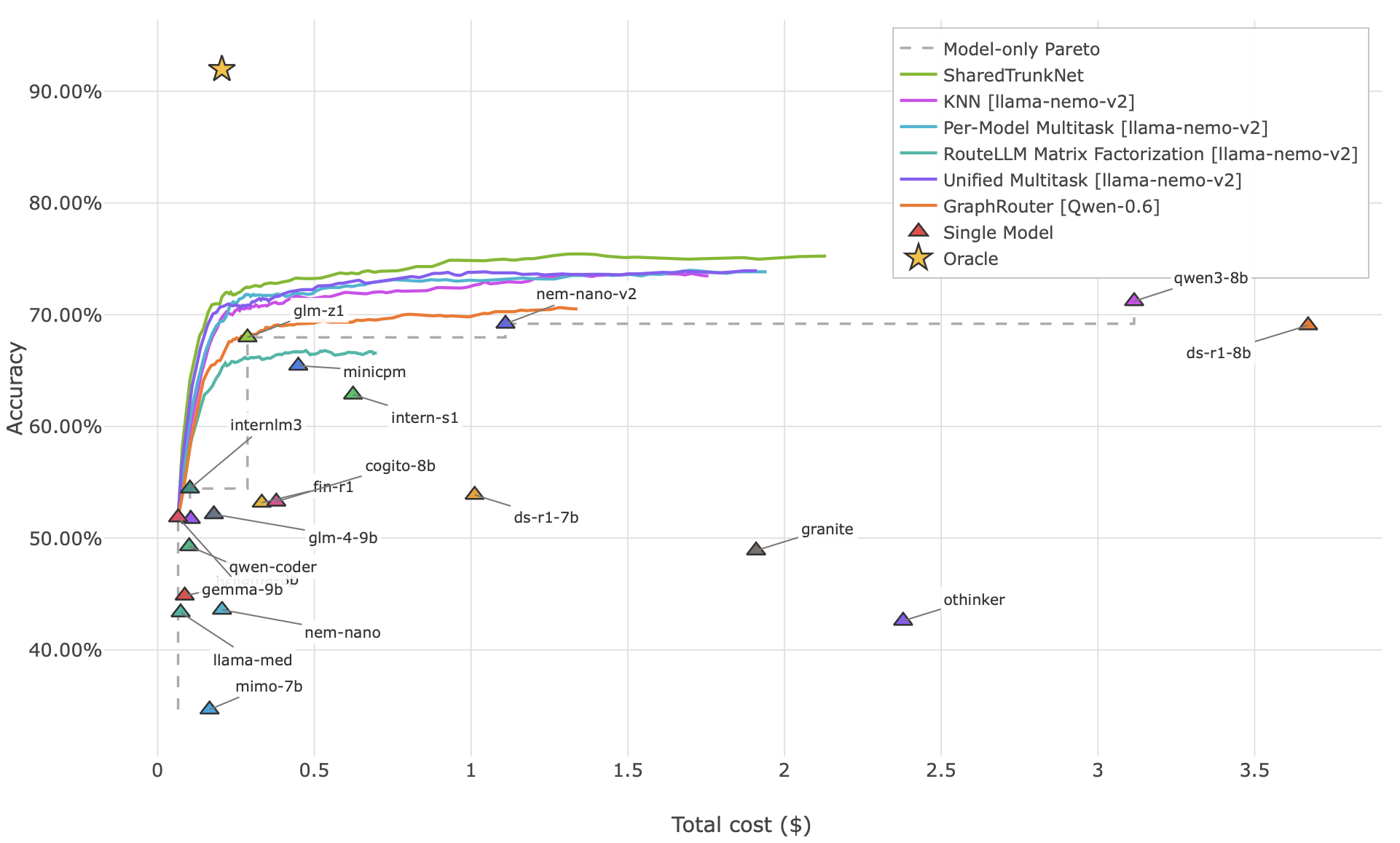}}}
    \raggedright
    \caption{Small pool: raw accuracy vs.\ total cost (\$).}
    \label{fig:small_raw}
  \end{minipage}
  \hfill
  \begin{minipage}[t]{0.38\linewidth}
    \centering
    {\setlength{\fboxsep}{0pt}\setlength{\fboxrule}{0.2pt}%
    \fbox{\includegraphics[width=\dimexpr\linewidth-2\fboxrule\relax]{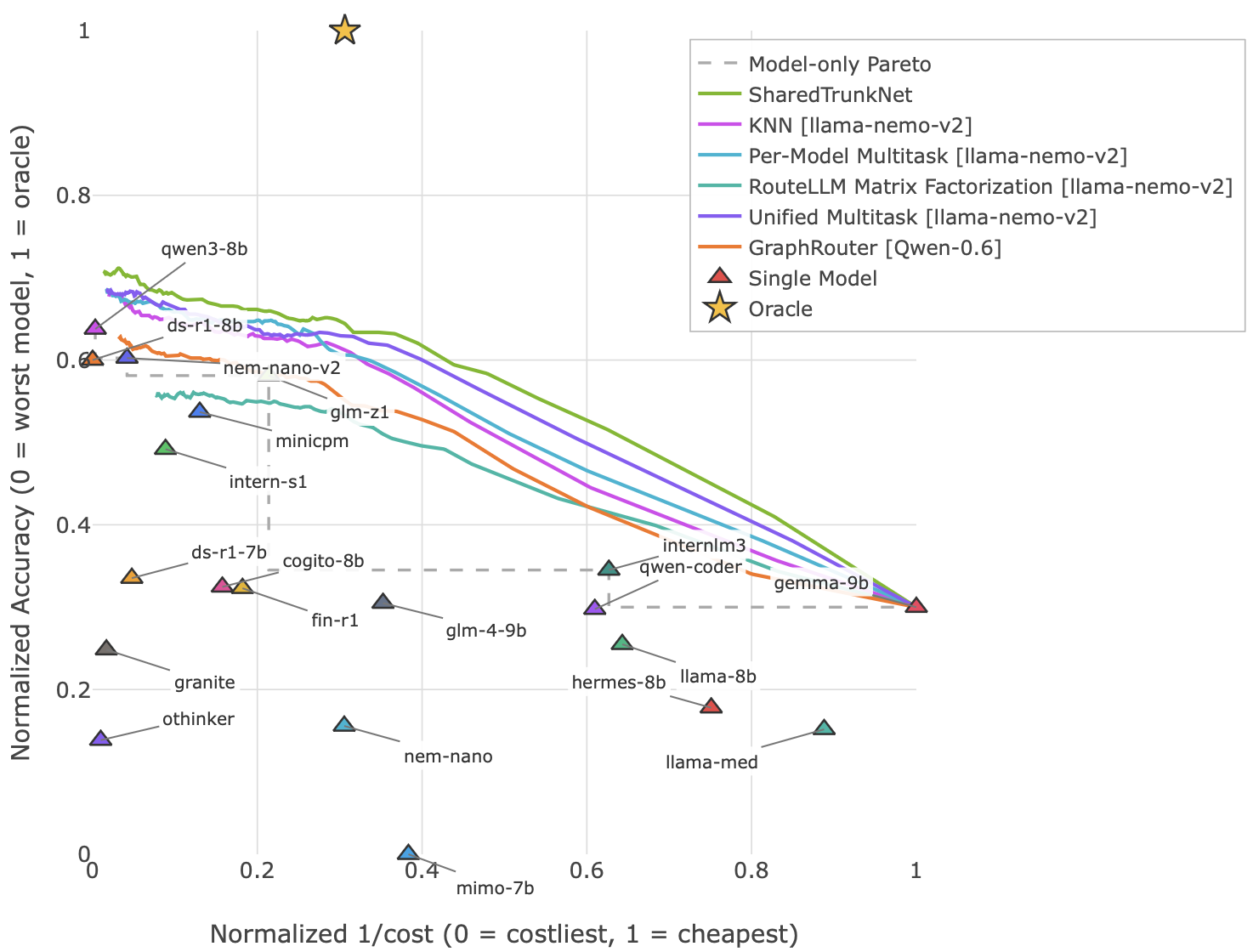}}}
    \caption{Small pool: normalized accuracy vs.\ normalized inverse cost.}
    \label{fig:small_norm}
  \end{minipage}
\end{figure}

Figures~\ref{fig:small_raw} and~\ref{fig:small_norm} show the raw and normalized accuracy--cost curves for the small model pool.

\subsection{Mixed Pool}

\begin{figure}[t]
  \centering
  \begin{minipage}[t]{0.58\linewidth}
    \centering
    {\setlength{\fboxsep}{0pt}\setlength{\fboxrule}{0.2pt}%
    \fbox{\includegraphics[width=\dimexpr\linewidth-2\fboxrule\relax]{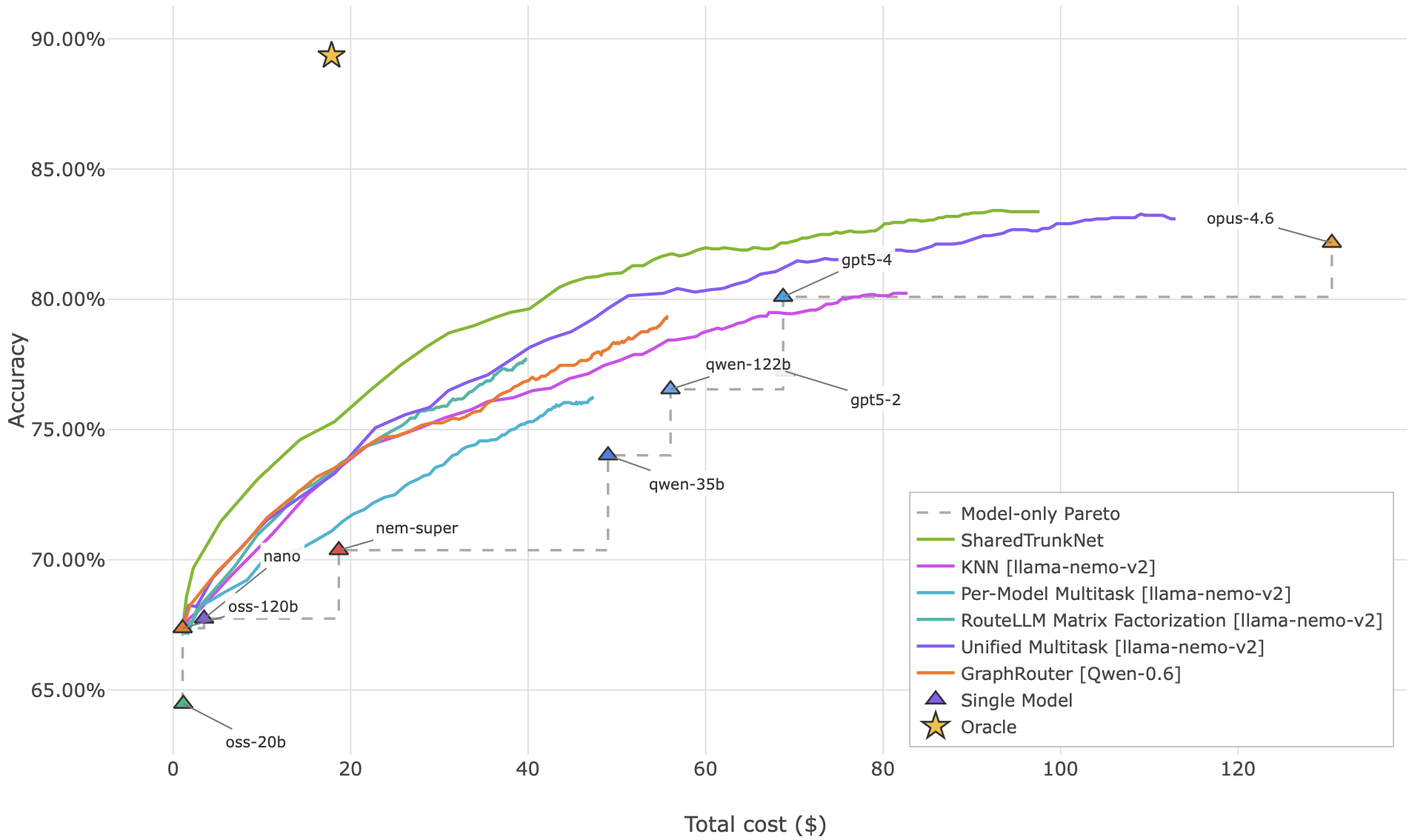}}}
    \raggedright
    \caption{Mixed pool: raw accuracy vs.\ total cost (\$).}
    \label{fig:mixed_raw}
  \end{minipage}
  \hfill
  \begin{minipage}[t]{0.38\linewidth}
    \centering
    {\setlength{\fboxsep}{0pt}\setlength{\fboxrule}{0.2pt}%
    \fbox{\includegraphics[width=\dimexpr\linewidth-2\fboxrule\relax]{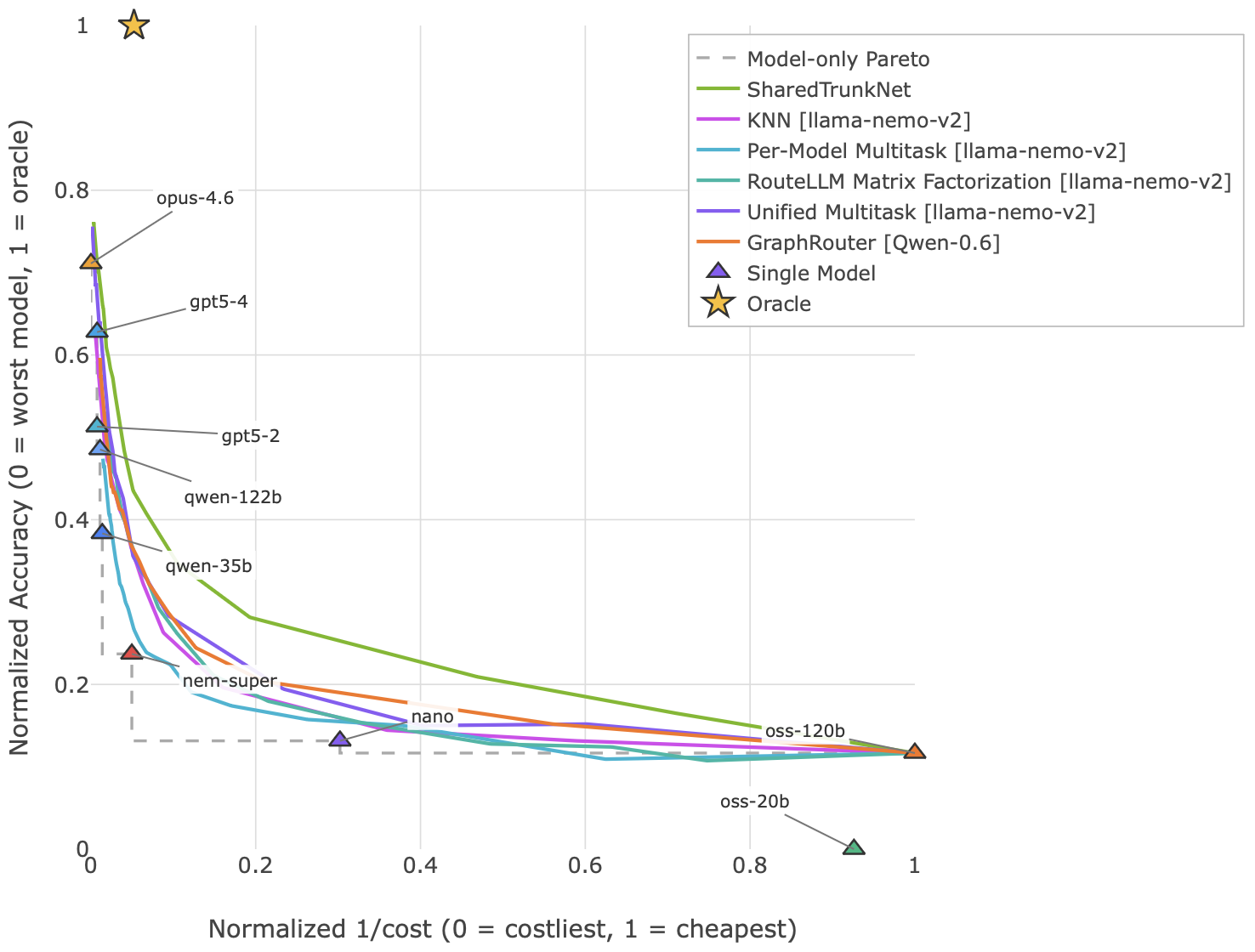}}}
    \caption{Mixed pool: normalized accuracy vs.\ normalized inverse cost.}
    \label{fig:mixed_norm}
  \end{minipage}
\end{figure}

Figures~\ref{fig:mixed_raw} and~\ref{fig:mixed_norm} show the raw and normalized accuracy--cost curves for the mixed model pool.

% ----------------------------------------------------------------
\clearpage
\section{Sample Distribution Across $\lambda$ Operating Points}
\label{sec:appendix_lambda}

Table~\ref{tab:lambda_operating_points} shows the sample distribution across $\lambda$ operating points for the mixed and frontier pools.

\begin{table}[t]
\begin{center}
\small
\setlength{\tabcolsep}{6pt}
\begin{tabular}{llrrrrr}
\toprule
\textbf{Pool} & \textbf{Step ($\lambda$)} & \textbf{N} & \textbf{Q1} & \textbf{Q2} & \textbf{Q3} & \textbf{Q4} \\
\midrule
Mixed    & $10^{-2}$ &  101 &    3 &    4 &   18 &   76 \\
Mixed    & $10^{-3}$ &  747 &   24 &   45 &  170 &  508 \\
Mixed    & $10^{-4}$ & 2198 &  239 &  397 &  728 &  834 \\
Mixed    & $10^{-5}$ & 3988 & 1243 &  910 &  940 &  895 \\
\midrule
Frontier & $10^{-2}$ &  101 &    8 &   10 &   22 &   61 \\
Frontier & $10^{-3}$ &  605 &   71 &   98 &  179 &  257 \\
Frontier & $10^{-4}$ & 1501 &  461 &  358 &  340 &  342 \\
Frontier & $10^{-5}$ & 1927 &  766 &  431 &  377 &  353 \\
\midrule
Small    & $10^{-2}$ &  101 &   41 &   22 &   23 &   15 \\
Small    & $10^{-3}$ &  550 &  216 &  119 &  115 &  100 \\
Small    & $10^{-4}$ &  805 &  390 &  146 &  143 &  126 \\
Small    & $10^{-5}$ & 1091 &  670 &  148 &  146 &  127 \\
\bottomrule
\end{tabular}
\end{center}
\caption{Sample distribution across $\lambda$ operating points for each pool. Each row shows
the total number of operating points \textbf{N} retained at a given threshold step size $\lambda$,
and how those points are distributed across cost-accuracy quartiles \textbf{Q1}--\textbf{Q4}
(Q1 = lowest cost, Q4 = highest cost). Finer steps yield more operating points with broader
quartile coverage, while coarser steps concentrate mass in higher-cost quartiles.}
\label{tab:lambda_operating_points}
\end{table}

\FloatBarrier

\end{document}